\documentclass[10pt,twocolumn,letterpaper]{article}

\usepackage{cvpr}              

\usepackage[dvipsnames]{xcolor}

\usepackage{makecell}
\usepackage{multirow}
\usepackage{colortbl}
\usepackage{bm}
\usepackage{indentfirst}
\usepackage{float}

\definecolor{cvprblue}{rgb}{0.21,0.49,0.74}
\usepackage[pagebackref,breaklinks,colorlinks,citecolor=cvprblue]{hyperref}
\usepackage[accsupp]{axessibility} 

\def\paperID{786} 
\def\confName{CVPR}
\def\confYear{2024}

\title{Coarse-to-Fine Latent Diffusion for Pose-Guided Person Image Synthesis}

\author{
Yanzuo Lu\textsuperscript{\rm 1} \and 
Manlin Zhang\textsuperscript{\rm 1} \and 
Andy J Ma\textsuperscript{\rm 1,\rm 2,\rm 3 \thanks{Corresponding author.}} \and 
Xiaohua Xie\textsuperscript{\rm 1,\rm 2,\rm 3} \and 
Jianhuang Lai\textsuperscript{\rm 1,\rm 2,\rm 3,\rm 4} \and
\textsuperscript{\rm 1}School of Computer Science and Engineering, Sun Yat-sen University, Guangzhou, China\\
\textsuperscript{\rm 2}Guangdong Province Key Laboratory of Information Security Technology, China\\
\textsuperscript{\rm 3}Key Laboratory of Machine Intelligence and Advanced Computing, Ministry of Education, China\\
\textsuperscript{\rm 4}Pazhou Lab (HuangPu), Guangzhou, China\\
{\tt\small \{luyz5, zhangmlin3\}@mail2.sysu.edu.cn, \{majh8, xiexiaoh6, stsljh\}@mail.sysu.edu.cn}
}

\begin{document}
\maketitle
\begin{abstract}
Diffusion model is a promising approach to image generation and has been employed for Pose-Guided Person Image Synthesis (PGPIS) with competitive performance.
While existing methods simply align the person appearance to the target pose, they are prone to overfitting due to the lack of a high-level semantic understanding on the source person image.
In this paper, we propose a novel Coarse-to-Fine Latent Diffusion (CFLD) method for PGPIS.
In the absence of image-caption pairs and textual prompts, we develop a novel training paradigm purely based on images to control the generation process of a pre-trained text-to-image diffusion model.
A perception-refined decoder is designed to progressively refine a set of learnable queries and extract semantic understanding of person images as a coarse-grained prompt.
This allows for the decoupling of fine-grained appearance and pose information controls at different stages, and thus circumventing the potential overfitting problem.
To generate more realistic texture details, a hybrid-granularity attention module is proposed to encode multi-scale fine-grained appearance features as bias terms to augment the coarse-grained prompt.
Both quantitative and qualitative experimental results on the DeepFashion benchmark demonstrate the superiority of our method over the state of the arts for PGPIS.
Code is available at \url{https://github.com/YanzuoLu/CFLD}.
\end{abstract}    
\section{Introduction}
\label{sec:introduction}

Pose-Guided Person Image Synthesis (PGPIS) aims to translate the source person image into a specific target pose while preserving the appearance as much as possible.
It has a wide range of applications, including film production, virtual reality, and fashion e-commerce.
Most existing methods along this line are developed based on Generative Adversarial Networks (GANs)~\cite{pg2,vunet,pose_gan,patn,diaf,gfla,xinggan,pise,spgnet,dptn,nted,disentangled,uvmap,liquid,adgan,casd}.
Nevertheless, the GAN-based approach may suffer from the instability of min-max training objective and difficulty in generating high-quality images in a single forward pass.


\begin{figure}[t]
    \centering
    \includegraphics[width=\linewidth]{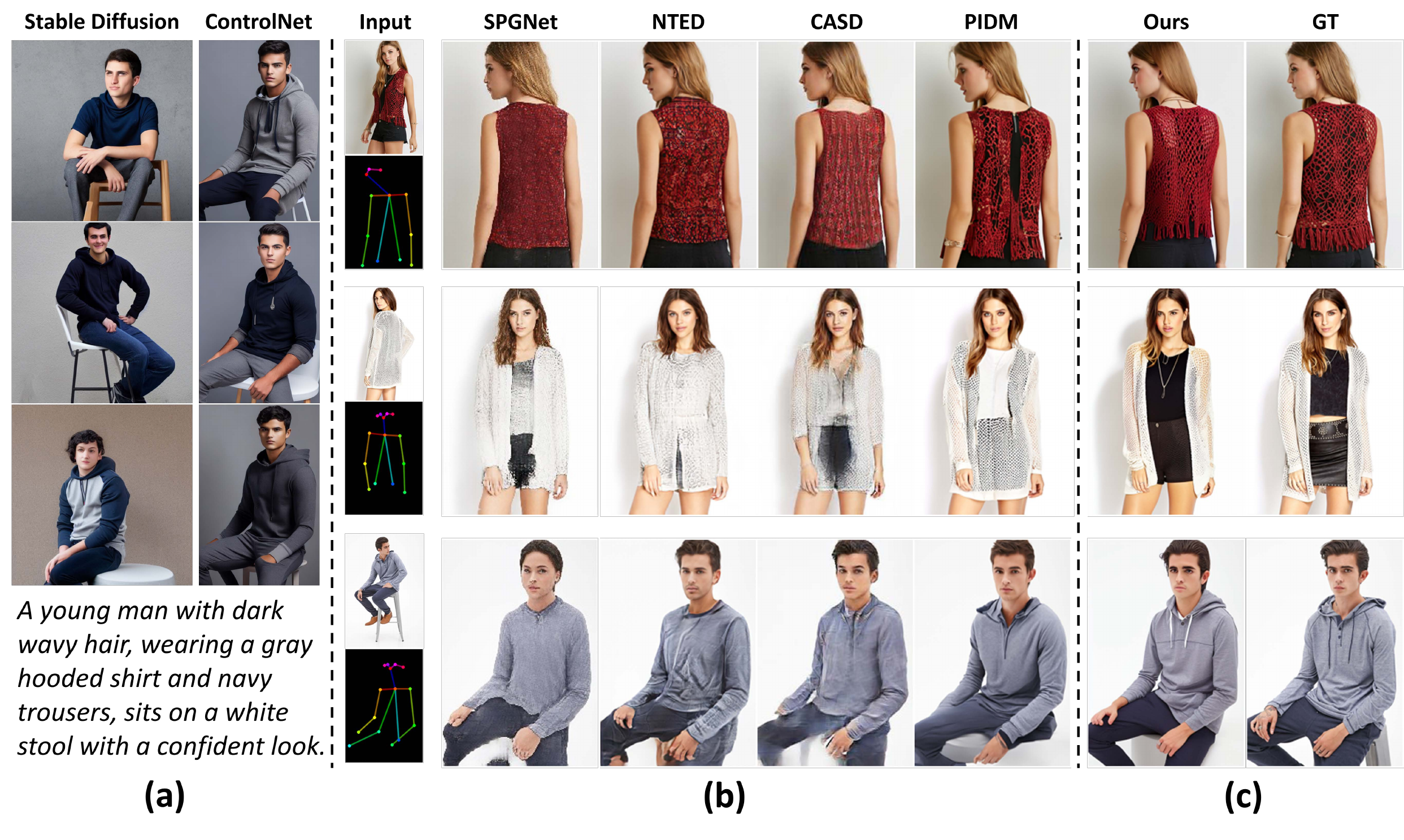}
    \vspace{-0.8cm}
    \caption{
        (a) The appearance of person images varies significantly given only a textual prompt for image generation by using Stable Diffusion~\cite{stable_diffusion} or ControlNet~\cite{controlnet} with OpenPose guidance~\cite{openpose}.
        (b) Simply aligning the source appearance to the target pose without a semantic understanding of person image can easily lead to overfitting, such that the generated images become distorted and unnatural.
        (c) Our method learns the coarse-grained prompt for a comprehensive perception of the source image and injects fine-grained appearance features as bias terms, thus generating high-quality images with better generalization performance.
    }
    \vspace{-0.3cm}
    \label{fig:intro}
\end{figure}

As a promising alternative to GANs for image generation, diffusion models synthesize more realistic images progressively from a series of denoising steps.
The recently prevailing text-to-image latent diffusion model, such as Stable Diffusion (SD)~\cite{stable_diffusion} may now generate compelling person images conditioned on a given textual prompt.
The appearance of the generated person can be determined by well-designed prompts~\cite{prompt1,prompt2} or prompt learning~\cite{coop,cocoop}.
With more reliable structural guidance~\cite{controlnet,t2i_adapter}, the synthesized person images can be further constrained to specific poses.
Though the text-to-image diffusion generates realistic images from textual prompts with high-level semantics, its training paradigm requires extensive image-caption pairs that are labor-expensive to collect for PGPIS.
More importantly, due to the differing information densities between language and vision~\cite{mae}, even the most detailed textual descriptions inevitably introduce ambiguity and may not accurately preserve the appearance as illustrated in \cref{fig:intro}(a).

More recently, several diffusion-based approaches have emerged for PGPIS.
A texture diffusion module is proposed by PIDM~\cite{pidm} to model the complex correspondence between the appearance of source image and the target pose.
Since the denoising process at the high-resolution pixel level is computationally expensive, PoCoLD~\cite{pocold} reduces both the training and inference costs by mapping pixels to low-dimensional latent spaces with a pre-trained Variational Autoencoder (VAE)~\cite{vae}.
In PoCoLD, the correspondence is further exploited by a pose-constrained attention module based on additional 3D Densepose~\cite{densepose} annotations.
While both the PIDM and PoCoLD generate more realistic texture details by aligning the source image to the target pose, they lack \textbf{a high-level semantic understanding of person images}.
Therefore, they are prone to overfitting and poor generalization performance when synthesizing exaggerated poses that are vastly different from the source image or rare in the training set.
As demonstrated in \cref{fig:intro}(b), the generated images become distorted and unnatural in these cases, which is in line with several GAN-based approaches.

In this work, we propose a novel Coarse-to-Fine Latent Diffusion (CFLD) method for PGPIS.
Our approach breaks the conventional training paradigm which leverages textual prompts to control the generation process of a pre-trained SD model.
Instead of conditioning on the human-generated signals, i.e. languages that are highly semantic and information-dense, we facilitate a coarse-to-fine appearance control method purely based on images.
To obtain the aforementioned semantic understanding specific to person images, we endeavor to decouple the fine-grained appearance and pose information controls at different stages by introducing a \textit{perception-refined decoder}.
The perception of the source person image is achieved by randomly initializing a set of learnable queries and progressively refining them in the following decoder blocks via cross-attention.
The decoder output serves as a coarse-grained prompt to describe the source image, focusing on the common semantics across different person images, e.g. human body parts and attributes such as age and gender.
Moreover, we design a \textit{hybrid-granularity attention} module to effectively encode multi-scale fine-grained appearance features as bias terms to augment the coarse-grained prompt.
In this way, the source image is able to be aligned with the target pose by supplementing only the necessary fine-grained details under the guidance of the coarse-grained prompt, thus achieving better generalization as illustrated in \cref{fig:intro}(c).

Our main contributions can be summarized as follows,
\begin{itemize}
\setlength{\itemsep}{0pt}
\setlength{\parsep}{0pt}
\setlength{\parskip}{0pt}
    \item We present a novel training paradigm in the absence of image-caption pairs to overcome the limitations when applying text-to-image diffusion to PGPIS. We propose a perception-refined decoder to extract semantic understanding of person images as a coarse-grained prompt.
    \item We formulate a new hybrid-granularity attention module to bias the coarse-grained prompt with fine-grained appearance features. Thus, the texture details of generated images are better controlled and become more realistic.
    \item We conduct extensive experiments on the DeepFashion~\cite{deepfashion} benchmark and achieve the state-of-the-art performance both quantitatively and qualitatively. User studies and ablations validate the effectiveness of our method.
\end{itemize}

\section{Related Work}
\label{sec:related_work}

\noindent\textbf{Pose-Guided Person Image Synthesis.}
Ma~\etal~\cite{pg2} first presents the task of pose-guided person image synthesis and refines the generated images in an adversarial manner.
To decouple the pose and appearance information, early approaches~\cite{vunet,disentangled} propose to learn pose-irrelevant features but fail to handle the complex texture details with vanilla convolutional neural networks.
To alleviate this problem, auxiliary information is introduced to improve the generation quality, such as parsing~\cite{adgan} and UV maps~\cite{uvmap}.
Recent approaches~\cite{patn,xinggan,diaf,gfla,liquid,nted} focus on modeling the spatial correspondence between pose and appearance, with the more frequent use of parsing maps~\cite{pise,spgnet,casd}.
PIDM~\cite{pidm} and PoCoLD~\cite{pocold} are developed based on diffusion models to prevent from the drawbacks in the generative adversarial networks, including the instability of min-max training objective and difficulty in synthesising high-resolution images.
Both of these two diffusion-based methods extend the idea of spatial correspondence to model the relation between the appearance of source image and target pose via the cross-attention mechanism.
We argue this leads to overfitting by simply aligning the source appearance to the target pose without a high-level semantic understanding of the person image. 
More concurrent works like MagicAnimate~\cite{magicanimate}, Animate Anyone~\cite{animate_anyone} and PCDMs~\cite{pcdms} require multi-stage and progressive fully fine-tuning, while our pipeline is more efficient and end-to-end by freezing most parameters.
And the training paradigm of IP-Adatper~\cite{ip_adapter} is heavily relying on image-text pairs which are not available for PGPIS task.


\begin{figure*}[t]
    \centering
    \includegraphics[width=0.95\linewidth]{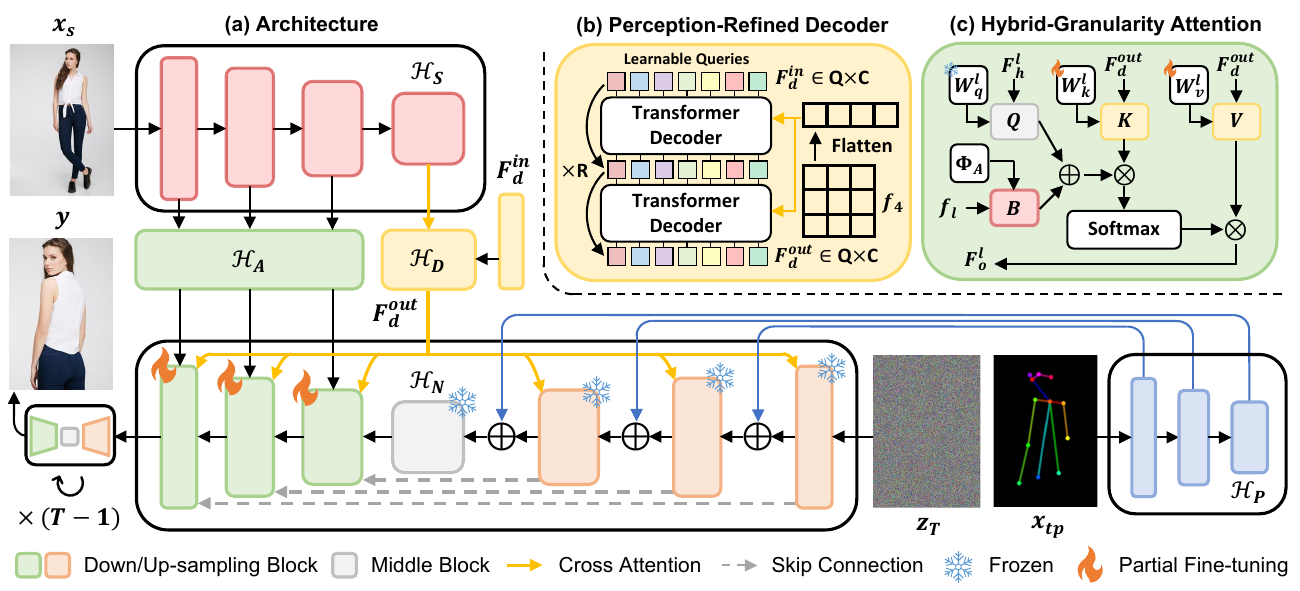}
    \vspace*{-0.15cm}
    \caption{
        (a) Architecture of our proposed Coarse-to-Fine Latent Diffusion (CFLD) method.
        For pose-guided latent diffusion, we incorporate a lightweight pose adapter $\mathcal{H}_P$ from \cite{t2i_adapter} to add its output feature maps to the end of each down-sampling block of the pre-trained UNet $\mathcal{H}_N$ for efficient structural guidance.
        To achieve a coarse-to-fine appearance control, we propose a perception-refined decoder $\mathcal{H}_D$ and hybrid-granularity attention module $\mathcal{H}_A$, both of which take the multi-scale feature maps from a source image encoder $\mathcal{H}_S$ as inputs.
        (b) The coarse-grained prompt is obtained by refining the learnable queries progressively in our proposed $\mathcal{H}_D$.
        (c) We encode the multi-scale fine-grained appearance features as bias terms in the up-sampling blocks for better texture details within $\mathcal{H}_A$.
    }
    \vspace*{-0.3cm}
    \label{fig:pipeline}
\end{figure*}

\noindent\textbf{Controllable Diffusion Models.}
Diffusion models have recently emerged and demonstrated their potential for high-resolution image synthesis.
The core idea is to start with a simple noise vector and gradually transform it into a high-quality image through multiple denoising iterations.
Beyond unconditional generation~\cite{ddpm,ddim,sde}, various methods have been introduced to incorporate user-supplied control signals into the generation process, enabling more controllable image generation.
For instance, \cite{dhariwal2021diffusion} introduces the usage of classifier gradients to condition on the generation, while \cite{cfg} proposes a classifier-free control mechanism employing a weighted summation of conditional and unconditional outputs for controllable synthesis.
Moreover, the Latent Diffusion Model (LDM) performs diffusion in the latent space and injects the conditioning signals via a specific encoder and cross-attention.
Building upon the pre-trained LDM like Stable Diffusion (SD)~\cite{stable_diffusion}, subsequent works have explored to bias the latent space by adding extra controls~\cite{controlnet,t2i_adapter}, as well as further to provide users with control over the generated content~\cite{tumanyan2023plug,hertz2022prompt}.
Rather than employing a high-level conditioning prompt throughout the generation, we design a coarse-to-fine conditioning process that adjusts the latent features at different stages within the UNet-based prediction network, providing better controllable pose-guided person image synthesis.

\section{Method}
\label{sec:method}




\subsection{Preliminary}
\label{sec:cldm}

Our method builds on top of the text-to-image latent diffusion model, i.e., Stable Diffusion (SD)~\cite{stable_diffusion} with high-quality image generation ability.
There are two main stages in the SD model: a Variational Autoencoder (VAE)~\cite{vae} that maps between raw-pixel space and low-dimensional latent space and an UNet-based prediction model~\cite{unet} for denoising diffusion image generation.
It follows the general idea of Denoising Diffusion Probabilistic Model (DDPM)~\cite{ddpm}, which formulates a forward diffusion process and a backward denoising process of $T=1000$ steps.
The diffusion process progressively adds random Gaussian noise $\epsilon \sim \mathcal{N}(0, \bm{I})$ to the initial latent $\bm{z}_0$, mapping it into noisy latents $\bm{z}_t$ at different timesteps $t\in [1,T]$, 
\begin{equation}
\label{eq:diffusion}
    \bm{z}_t=\sqrt{\bar\alpha_t}\bm{z}_0 + \sqrt{1-\bar\alpha_t}\epsilon,
\end{equation}
where $\bar\alpha_1,\bar\alpha_2,...,\bar\alpha_T$ are derived from a fixed variance schedule.
The denoising process learns the UNet $\epsilon_\theta(\bm{z}_t,t,\bm{c})$ to predict the noise and reverse this mapping, where $\bm{c}$ is the conditional embedding output by e.g. the CLIP~\cite{clip} text encoder in~\cite{stable_diffusion}.
The optimization can be formulated as,
\begin{equation}
\label{eq:mse}
    \mathcal{L}_{mse} = \mathbb{E}_{\bm{z}_0,\bm{c},\epsilon,t}\left[\|\epsilon-\epsilon_\theta(\bm{z}_t,t,\bm{c})\|_2^2\right].
\end{equation}



\subsection{Coarse-to-Fine Latent Diffusion}

\noindent\textbf{Architecture and Overview.}
\cref{fig:pipeline}(a) shows the architecture of our proposed method.
For concise illustration, we omit the encoder $\mathcal{E}$ and decoder $\mathcal{D}$ of the VAE~\cite{vae} model in this figure.
In the training phase, we are given sets of the source image $\bm{x}_s$, source pose $\bm{x}_{sp}$, target pose $\bm{x}_{tp}$, and ground-truth image $\bm{x}_g$.
The source image passes through an image encoder $\mathcal{H}_S$ (e.g. swin transformer~\cite{swin_transformer}), from which we extract a stack of multi-scale feature maps $\bm{F}_s=[\bm{f}_1,\bm{f}_2,\bm{f}_3,\bm{f}_4]$ for a coarse-to-fine \textit{appearance control}.
The coarse-grained prompts are learned by our Perception-Refined Decoder (PRD) $\mathcal{H}_D$ and serve as conditional embeddings in both down-sampling and up-sampling blocks of the UNet $\mathcal{H}_N$.
While the down-sampling block in $\mathcal{H}_N$ remains intact in our method, we reformulate the up-sampling block with our Hybrid-Granularity Attention module (HGA) $\mathcal{H}_A$ to bias the coarse-grained prompt with fine-grained appearance features for more realistic textures.
More details about $\mathcal{H}_D$ and $\mathcal{H}_A$ will be presented later.

For efficient \textit{pose control}, we adopt a lightweight pose adapter $\mathcal{H}_P$ that consists of several ResNet blocks~\cite{resnet}.
The output feature maps of $\mathcal{H}_P$ are added directly to the end of each down-sampling block as in \cite{t2i_adapter}.
This requires no additional fine-tuning and explicitly decouples the fine-grained appearance and pose information controls.
At different scales of down-sampling, the pose information is only aligned with the same coarse-grained prompts given by our PRD as conditional embeddings, rather than the different multi-scale fine-grained appearance features in the common practice~\cite{pidm,pocold}.
In this way, the HGA module learns all the pose-irrelevant texture details at the up-sampling stage and is not prone to overfitting.
Denote the initial latent state for the ground-truth image as $\bm{z}_0=\mathcal{E}(\bm{x}_g)$.
The MSE loss in \cref{eq:mse} is thus rewritten as,
\begin{equation}
    \mathcal{L}_{mse} = \mathbb{E}_{\bm{z}_0,\bm{x}_s,\bm{x}_{tp},\epsilon,t}\left[\|\epsilon-\epsilon_\theta(\bm{z}_t,t,\bm{x}_s,\bm{x}_{tp})\|_2^2\right].
\end{equation}



\noindent\textbf{Perception-Refined Decoder.}
Instead of utilizing multi-scale appearance features as conditional embeddings as in the existing diffusion-based approaches~\cite{pidm,pocold}, we propose to decouple the controls from the fine-grained appearance and pose information at different stages.
Thus we design a Perception-Refined Decoder (PRD) to extract semantic understanding of person images as a coarse-grained prompt, given the flattened last-scale output $\bm{f}_4$ from $\mathcal{H}_S$ as illustrated in \cref{fig:pipeline}(b).
By revisiting how people perceive a person image, we find several common characteristics, i.e., human body parts, age, gender, hairstyle, clothing, and so on, as demonstrated in \cref{fig:intro}(a).
This inspires us to maintain a set of learnable queries $\bm{F}_{d}^{in}\in\mathbb{R}^{Q\times D}$ representing different semantics of person images.
They are randomly initialized and progressively refined with the standard transformer decoders~\cite{transformer}.
The source image conditioning $\bm{f}_4$ interacts via the cross-attention module at each decoder block.
After $R$ blocks of refinement, we obtain the coarse-grained prompt $\bm{F}_{d}^{out}$, which serves as the conditional embedding and inputs to both down-sampling and up-sampling in $\mathcal{H}_N$.

\noindent\textbf{Hybrid-Granularity Attention.}
To precisely control the texture details of generated images, we introduce the Hybrid-Granularity Attention module (HGA) that is embedded in different scales $(l\in\{1,2,3\})$ of up-sampling blocks in $\mathcal{H}_N$, where we refer $\bm{F}_h^l,\bm{F}_o^l$ to its input and output.
Given the multi-scale feature maps $\bm{f}_l$ of the source image from $\mathcal{H}_S$, the HGA module aims to compensate for the missing necessary details in the coarse-grained prompts.
To achieve this, we formulate the HGA module that naturally follows a coarse-to-fine learning curriculum.

Specifically, we propose to inject multi-scale texture details by biasing the queries of cross-attention in the up-sampling blocks as shown in~\cref{fig:pipeline}(c), i.e.,
\begin{equation}
\begin{gathered}
    \bm{Q}=\bm{W}_q^l\bm{F}_h^l, \quad\bm{K}=\bm{W}_k^l \bm{F}_d^{out},\quad\bm{V}=\bm{W}_v^l \bm{F}_d^{out}, \\
    \bm{B}=\bm{\phi}_A(\bm{f}_l), \quad\bm{F}_o^l=\text{softmax}(\frac{(\bm{Q}+\bm{B})\bm{K}^T}{\sqrt{d}})\bm{V}, 
\end{gathered}
\end{equation}
where $\bm{W}_q^l,\bm{W}_k^l,\bm{W}_v^l$ are specific projection layers for the $l$-th scale up-sampling block of dimension $d$.
$\bm{\phi}_A$ is a fine-grained appearance encoder that mainly consists of $K$ transformer layers with a zero convolution~\cite{controlnet} added in the beginning and the end.
The zero convolution is a standard $1\times1$ convolution layer with both weight and bias initialized as zeros.
It keeps the gradient of $\mathcal{H}_A$ back to $\mathcal{H}_S$ small enough in the early stage of training, so that the image encoder $\mathcal{H}_S$ and the more easily converged perception-refined decoder $\mathcal{H}_D$ can focus on learning to provide a high-level semantic understanding compatible with the pre-trained SD model.
Since we have decoupled the controls of the fine-grained appearance and pose information at different stages, the target pose can be well controlled without overfitting during the down-sampling process.
Therefore, such a design encourages the HGA module to slowly fill in more fine-grained textures to better align the generation with the source image during training.
Note that $\bm{W}_k^l,\bm{W}_v^l$ in the up-sampling blocks are trainable parameters.
They are the only trainable parameters of the entire $\mathcal{H}_N$, which accounts for only \textbf{1.2\%} of all the parameters in the pre-trained SD model.

\subsection{Optimization}

To assist the source-to-target pose translation, we follow the insights in~\cite{dptn} to conduct source-to-source self-reconstruction for training. The reconstruction loss is,
\begin{equation}
    \mathcal{L}_{rec} = \mathbb{E}_{\bm{{z}}_0,\bm{x}_s,\bm{x}_{sp},\epsilon,t}\left[\|\epsilon-\epsilon_\theta(\bm{{z}}_t,t,\bm{x}_s,\bm{x}_{sp})\|_2^2\right],
\end{equation}
where $\bm{{z}}_0=\mathcal{E}(\bm{x}_s)$ and $\bm{{z}}_t$ is the noisy latent mapped from $\bm{{z}}_0$ at timestep $t$.
The overall objective is written as,
\begin{equation}
    \mathcal{L}_{overall} = \mathcal{L}_{mse} + \mathcal{L}_{rec}.
\end{equation}
Moreover, we adopt the cubic function $t=(1-(\frac{t}{T})^3)\times T,\ t\in\text{Uniform}(1,T)$ for the distribution of timestep $t$.
It increases the probability of $t$ falling in the early sampling stage and strengthens the guidance, which helps to converge faster and shorten the training time.

\noindent\textbf{Sampling.}
Once the conditional latent diffusion model is learned, the inference can be performed and starts by sampling a random Gaussian noise $\bm{z}_T \sim \mathcal{N}(0, \bm{I})$.
The predicted latent $\tilde{\bm{z}}_0$ is obtained by reversing the schedule in \cref{eq:diffusion} using the denoising network $\epsilon_t$ at each timestep $t\in[1,T]$.
We adopt the cumulative classifier-free guidance~\cite{pocold,cfg,cumulative_cfg} to strengthen both the source appearance and target pose guidance, i.e.,
\begin{equation}\label{eq:cfg}
\begin{aligned}
    \epsilon_t = &\ \epsilon_\theta(\bm{z}_t,t,\varnothing,\varnothing) \\
    & + w_{\text{pose}} (\epsilon_\theta(\bm{z}_t,t,\varnothing,\bm{x}_{tp}) - \epsilon_\theta(\bm{z}_t,t,\varnothing,\varnothing)) \\
    & + w_{\text{app}} (\epsilon_\theta(\bm{z}_t,t,\bm{x}_s,\bm{x}_{tp}) - \epsilon_\theta(\bm{z}_t,t,\varnothing,\bm{x}_{tp})).
\end{aligned}
\end{equation}
When the source image $\bm{x}_s$ is missing, we use learnable vectors as the conditional embeddings.
The learnable vectors are trained with a probability of $\eta$\% to drop both $\bm{x}_s$ and $\bm{x}_p$ during training.
The outputs of the pose adapter $\mathcal{H}_P$ will be set to all zeros if the target pose $\bm{x}_{tp}$ is missing.
We use the DDPM scheduler~\cite{ddpm} with 50 steps as the same as in~\cite{pocold,pidm}.
Finally, the generated image is obtained by the VAE decoder $\bm{y}=\mathcal{D}(\tilde{\bm{z}}_0)$.

\section{Experiments}
\label{sec:experiment}

\begin{table}[t]
    \centering
    \footnotesize
    \begin{tabular}{llcc}
        \toprule
        \textbf{Component} & \multicolumn{2}{l}{\textbf{Default}} & \makecell[c]{\textbf{Trainable}\\\textbf{Params.}} \\
        \midrule
        $\mathcal{H}_S$ & \multicolumn{2}{l}{Swin-B~\cite{swin_transformer}} & 87.0M \\
        $\mathcal{H}_A$ & \multicolumn{2}{l}{$K=4$} & 22.5M \\
        $\mathcal{H}_D$ & \multicolumn{2}{l}{$R=8$, $Q=16$, $C=768$} & 97.7M \\
        $\mathcal{H}_P$ & \multicolumn{2}{l}{Adapter \cite{t2i_adapter}} & 30.6M \\
        $\mathcal{H}_N$ & \multicolumn{2}{l}{up-sampling $\bm{W}_k^l,\bm{W}_v^l$} & 10.3M \\
        \midrule
        \textbf{Method} & \makecell[l]{\textbf{Pose Info. \&}\\\textbf{Annotation}} & \makecell[c]{\textbf{Training}\\\textbf{Epochs}} & \makecell[c]{\textbf{Trainable}\\\textbf{Params.}} \\
        \midrule
        PIDM~\cite{pidm} & 2D OpenPose~\cite{openpose} & 300 & 688.0M \\
        PoCoLD~\cite{pocold} & 3D DensePose~\cite{densepose} & 100 & 395.9M \\
        \textbf{CFLD (Ours)} & 2D OpenPose~\cite{openpose} & 100 & \textbf{248.2M} \\
        \bottomrule
    \end{tabular}
    \caption{The default settings and the number of trainable parameters in each component of our method and comparison with other diffusion-based methods.}
    \vspace{-0.3cm}
    \label{tab:hyperparameter}
\end{table}
\begin{table}[t]
    \centering
    \footnotesize
    \begin{tabular}{m{1.8cm}m{1cm}*{4}{m{0.7cm}}}
        \toprule
        \makebox[1.8cm][c]{\textbf{Method}} & \makebox[1cm][l]{\textbf{Venue}} &\makebox[0.7cm][c]{\textbf{FID$\downarrow$}} & \makebox[0.7cm][c]{\textbf{LPIPS$\downarrow$}} & \makebox[0.7cm][c]{\textbf{SSIM$\uparrow$}} & \makebox[0.7cm][c]{\textbf{PSNR$\uparrow$}}\\
        \midrule

        \rowcolor[gray]{0.9} \multicolumn{6}{l}{\textit{Evaluate on 256$\times$176 resolution}} \\
        
        \makebox[1.8cm][l]{PATN~\cite{patn}} & \makebox[1cm][r]{CVPR 19'} & \makebox[0.7cm][r]{20.728} & \makebox[0.7cm][c]{0.2533} & \makebox[0.7cm][c]{0.6714} & \makebox[0.7cm][c]{-} \\
        
        \makebox[1.8cm][l]{ADGAN~\cite{adgan}} & \makebox[1cm][r]{CVPR 20'} & \makebox[0.7cm][r]{14.540} & \makebox[0.7cm][c]{0.2255} & \makebox[0.7cm][c]{0.6735} & \makebox[0.7cm][c]{-} \\
        
        \makebox[1.8cm][l]{GFLA~\cite{gfla}} & \makebox[1cm][r]{CVPR 20'} & \makebox[0.7cm][r]{9.827} & \makebox[0.7cm][c]{0.1878} & \makebox[0.7cm][c]{0.7082} & \makebox[0.7cm][c]{-} \\
        
        \makebox[1.8cm][l]{PISE~\cite{pise}} & \makebox[1cm][r]{CVPR 21'} & \makebox[0.7cm][r]{11.518} & \makebox[0.7cm][c]{0.2244} & \makebox[0.7cm][c]{0.6537} & \makebox[0.7cm][c]{-} \\
        
        \makebox[1.8cm][l]{SPGNet$^\dagger$~\cite{spgnet}} & \makebox[1cm][r]{CVPR 21'} & \makebox[0.7cm][r]{16.184} & \makebox[0.7cm][c]{0.2256} & \makebox[0.7cm][c]{0.6965} & \makebox[0.7cm][c]{17.222} \\
        
        \makebox[1.8cm][l]{DPTN$^\dagger$~\cite{dptn}} & \makebox[1cm][r]{CVPR 22'} & \makebox[0.7cm][r]{17.419} & \makebox[0.7cm][c]{0.2093} & \makebox[0.7cm][c]{0.6975} & \makebox[0.7cm][c]{17.811} \\
        
        \makebox[1.8cm][l]{NTED$^\dagger$~\cite{nted}} & \makebox[1cm][r]{CVPR 22'} & \makebox[0.7cm][r]{8.517} & \makebox[0.7cm][c]{0.1770} & \makebox[0.7cm][c]{0.7156} & \makebox[0.7cm][c]{17.740} \\
        
        \makebox[1.8cm][l]{CASD$^\dagger$~\cite{casd}} & \makebox[1cm][r]{ECCV 22'} & \makebox[0.7cm][r]{13.137} & \makebox[0.7cm][c]{0.1781} & \makebox[0.7cm][c]{0.7224} & \makebox[0.7cm][c]{\underline{17.880}} \\
        
        \makebox[1.8cm][l]{PIDM$^\dagger$~\cite{pidm}} & \makebox[1cm][r]{CVPR 23'} & \makebox[0.7cm][r]{\underline{6.812}} & \makebox[0.7cm][c]{0.2006} & \makebox[0.7cm][c]{0.6621} & \makebox[0.7cm][c]{15.630} \\

        \makebox[1.8cm][l]{\textcolor{gray}{PIDM$^\ddagger$~\cite{pidm}}} & \makebox[1cm][r]{\textcolor{gray}{CVPR 23'}} & \makebox[0.7cm][r]{\textcolor{gray}{6.440}} & \makebox[0.7cm][c]{\textcolor{gray}{0.1686}} & \makebox[0.7cm][c]{\textcolor{gray}{0.7109}} & \makebox[0.7cm][c]{\textcolor{gray}{17.399}} \\
        
        \makebox[1.8cm][l]{PoCoLD~\cite{pocold}} & \makebox[1cm][r]{ICCV 23'} & \makebox[0.7cm][r]{8.067} & \makebox[0.7cm][c]{\underline{0.1642}} & \makebox[0.7cm][c]{\underline{0.7310}} & \makebox[0.7cm][c]{-} \\

        \multicolumn{2}{l}{\textbf{CFLD (Ours)}} & \makebox[0.7cm][r]{\textbf{6.804}} & \makebox[0.7cm][c]{\textbf{0.1519}} & \makebox[0.7cm][c]{\textbf{0.7378}} & \makebox[0.7cm][c]{\textbf{18.235}} \\
        
        \multicolumn{2}{l}{\textcolor{gray}{VAE Reconstructed}} & \makebox[0.7cm][r]{\textcolor{gray}{7.967}} & \makebox[0.7cm][c]{\textcolor{gray}{0.0104}} & \makebox[0.7cm][c]{\textcolor{gray}{0.9660}} & \makebox[0.7cm][c]{\textcolor{gray}{33.515}} \\
        
        \multicolumn{2}{l}{\textcolor{gray}{Ground Truth}} & \makebox[0.7cm][r]{\textcolor{gray}{7.847}} & \makebox[0.7cm][c]{\textcolor{gray}{0.0000}} & \makebox[0.7cm][c]{\textcolor{gray}{1.0000}} & \makebox[0.7cm][c]{\textcolor{gray}{$+\infty$}} \\

        \midrule

        \rowcolor[gray]{0.9} \multicolumn{6}{l}{\textit{Evaluate on 512$\times$352 resolution}} \\

        \makebox[1.8cm][l]{CoCosNet2~\cite{cocosnetv2}} & \makebox[1cm][r]{CVPR 21'} & \makebox[0.7cm][r]{13.325} & \makebox[0.7cm][c]{0.2265} & \makebox[0.7cm][c]{0.7236} & \makebox[0.7cm][c]{-} \\
        
        \makebox[1.8cm][l]{NTED$^\dagger$~\cite{nted}} & \makebox[1cm][r]{CVPR 22'} & \makebox[0.7cm][r]{\underline{7.645}} & \makebox[0.7cm][c]{0.1999} & \makebox[0.7cm][c]{0.7359} & \makebox[0.7cm][c]{\underline{17.385}} \\
        
        \makebox[1.8cm][l]{PoCoLD~\cite{pocold}} & \makebox[1cm][r]{ICCV 23'} & \makebox[0.7cm][r]{8.416} & \makebox[0.7cm][c]{\underline{0.1920}} & \makebox[0.7cm][c]{\underline{0.7430}} & \makebox[0.7cm][c]{-} \\

        \multicolumn{2}{l}{\textbf{CFLD (Ours)}} & \makebox[0.7cm][r]{\textbf{7.149}} & \makebox[0.7cm][c]{\textbf{0.1819}} & \makebox[0.7cm][c]{\textbf{0.7478}} & \makebox[0.7cm][c]{\textbf{17.645}} \\
        
        \multicolumn{2}{l}{\textcolor{gray}{VAE Reconstructed}} & \makebox[0.7cm][r]{\textcolor{gray}{8.187}} & \makebox[0.7cm][c]{\textcolor{gray}{0.0217}} & \makebox[0.7cm][c]{\textcolor{gray}{0.9231}} & \makebox[0.7cm][c]{\textcolor{gray}{30.214}} \\
        
        \multicolumn{2}{l}{\textcolor{gray}{Ground Truth}} & \makebox[0.7cm][r]{\textcolor{gray}{8.010}} & \makebox[0.7cm][c]{\textcolor{gray}{0.0000}} & \makebox[0.7cm][c]{\textcolor{gray}{1.0000}} & \makebox[0.7cm][c]{\textcolor{gray}{$+\infty$}} \\

        \bottomrule
    \end{tabular}
    \caption{Quantitative comparisons with the state of the arts in terms of image quality. $^\dagger$ We strictly follow the evaluation implementation in NTED~\cite{nted} and reproduce these results. 
    $^\ddagger$ Results are obtained using the generated images released by the authors.
    }
    \label{tab:quantitative}
    \vspace{-0.3cm}
\end{table}
\begin{figure*}[t]
    \centering
    \includegraphics[width=\linewidth]{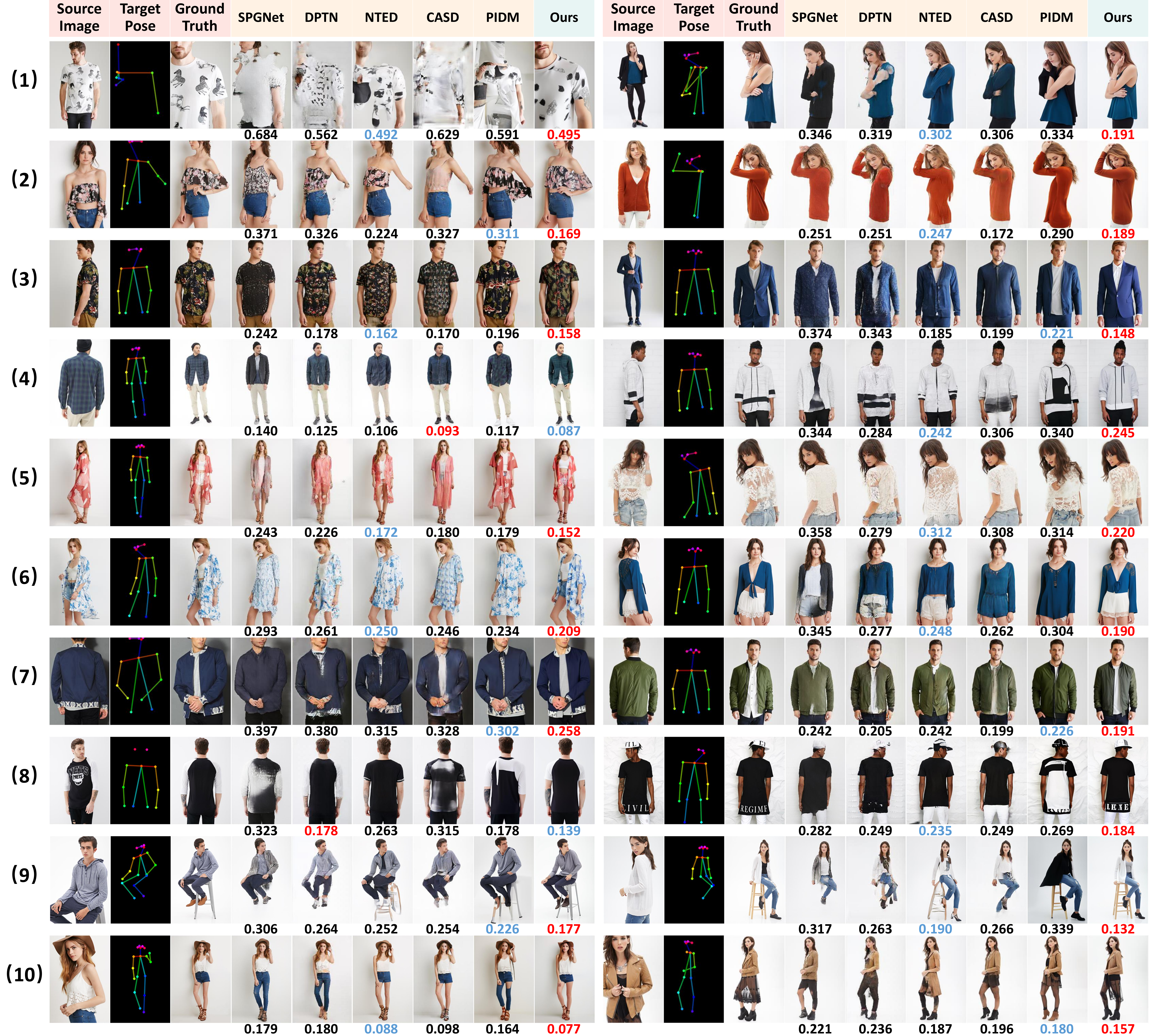}
    \vspace*{-0.5cm}
    \caption{
        Qualitative comparisons with state-of-the-arts. To clarify the relation between objective and subjective metrics, we demonstrate the LPIPS measures and label the images with the first and second highest votes from user opinions as \textcolor{red}{red} and \textcolor{blue}{blue} respectively.
    }
    \vspace*{-0.3cm}
    \label{fig:qualitative}
\end{figure*}

\subsection{Setup}

\noindent\textbf{Dataset.}
We follow \cite{pocold,nted} to conduct experiments on the In-Shop Clothes Retrieval benchmark of DeepFashion~\cite{deepfashion} and evaluate on both the 256$\times$176 and 512$\times$352 resolutions.
This dataset consists of 52,712 high-resolution person images in the fashion domain.
The dataset split is the same as in PATN~\cite{patn}, where 101,966 and 8,570 non-overlapping pairs are selected for training and testing, respectively.

\noindent\textbf{Objective metrics.}
We use four different metrics to evaluate the generated images quantitatively, including FID~\cite{fid}, LPIPS~\cite{lpips}, SSIM~\cite{ssim} and PSNR.
Both FID and LPIPS are based on deep features. The Fréchet Inception Distance (FID) calculates the Wasserstein-2 distance~\cite{wasserstein} between the distributions of generated and real images using Inception-v3~\cite{inception} features, and the Learned Perceptual Image Patch Similarity (LPIPS) leverages a network trained on human judgments to measure reconstruction accuracy in the perceptual domain.
As for the Structural Similarity Index Measure (SSIM) and Peak Signal to Noise Ratio (PSNR), they quantify the similarity between generated images and ground truths at the pixel level.

\noindent\textbf{Subjective metrics.}
In addition to the objective metrics, we follow \cite{pocold} to use the Jab~\cite{spgnet,casd,pidm} metric in our user study to calculate the percentage of generated images that were considered the best among all methods~\cite{pidm,casd,nted,dptn,spgnet}.
Moreover, in order to measure the similarity between the generated images and real data, we quantify the R2G and G2R metrics as many early approaches did~\cite{pg2,pose_gan,patn}.
R2G represents the percentage of real images considered as generated and G2R represents the percentage of generated images considered as real by humans.

\noindent\textbf{Implementation details.}
Our method is implemented with PyTorch~\cite{pytorch} and HuggingFace Diffusers~\cite{diffusers} on top of the Stable Diffusion~\cite{stable_diffusion} with the version of 1.5.
The source image is resized to 256$\times$256 and the source image encoder $\mathcal{H}_S$ is a standard Swin-B~\cite{swin_transformer} pretrained on ImageNet~\cite{imagenet}.
The default settings and the number of trainable parameters in each component are summarized in \cref{tab:hyperparameter}.
We train for 100 epochs using the Adam~\cite{adam} optimizer with a base learning rate of 5e-7 scaled by the total batch size.
The learning rate undergoes a linear warmup during the first 1,000 steps and is multiplied by 0.1 at 50 epochs.
For classifier-free guidance, we set $w_\text{pose}$ and $w_\text{id}$ to 2.0 during sampling, and drop the condition $\bm{x}_s$ and $\bm{x}_p$ with a probability of $\eta=20$(\%) during training.

\subsection{Quantitative Comparison}
We quantitatively compare our method with both GAN-based and diffusion-based state-of-the-art approaches in terms of objective metrics.
The evaluation is performed on both 256$\times$176 and 512$\times$352 resolutions as the same as in~\cite{nted,pocold}.
As shown in \cref{tab:quantitative}, our method significantly outperforms the state-of-the-art across all metrics on both resolutions.
In particular, compared to the other two diffusion-based methods~\cite{pidm,pocold} in \cref{tab:hyperparameter}, we achieve better reconstruction with simpler 2D-only pose annotations and fewer trainable parameters.
The metrics for VAE~\cite{vae} reconstructions and the ground truths are also provided for reference.
It is worth noting that the results we obtain by running with the checkpoint provided by PIDM~\cite{pidm} suffer from severe overfitting, resulting in a large gap between the quantitative results of provided images and those from the checkpoint.

\subsection{Qualitative Comparison}
In \cref{fig:qualitative}, we present a comprehensive visual comparison with recent approaches that are publicly available and reproducible, including SPGNet~\cite{spgnet}, DPTN~\cite{dptn}, NTED~\cite{nted}, CASD~\cite{casd} and PIDM~\cite{pidm}.
Our observations can be summarized as follows.
\textbf{(1)} Both GAN-based and diffusion-based methods suffer from overfitting the human poses.
When generating some target poses that are extreme or not common in the training set, existing methods show severe distortions as demonstrated in \verb|rows 1-2|.
Since we decouples the controls of fine-grained appearance and pose information, our method circumvents the potential overfitting problem and always generates a reasonable pose with the conditioning coarse-grained prompt and fine-grained appearance bias.
\textbf{(2)} For source images in \verb|rows 3-6| with more complex clothing, our generated images better preserve the textures details while aligning with the target pose thanks to the robust coarse-to-fine learning curriculum of hybrid-granularity attention module.
For other methods, although they match in color, the clothes either exhibit blurring and distortion (SPGNet, DPTN, and CASD) or are spliced unnaturally in texture, creating a large gap from the source image (NTED and PIDM).
\textbf{(3)} As for cases where the target pose requires visualization of areas invisible in the source image, our method exhibits strong understanding and generalization capabilities.
With a semantic understanding of the source image provided by the perception-refined decoder, our method is aware of what should be predicted when the person turns around or sits down as illustrated in \verb|rows 7-10|, such as different patterns on the front and back of clothes, the sitting chair, and lower body wear.


\begin{figure}[t]
    \centering
    \vspace{-0.2cm}
    \includegraphics[width=\linewidth]{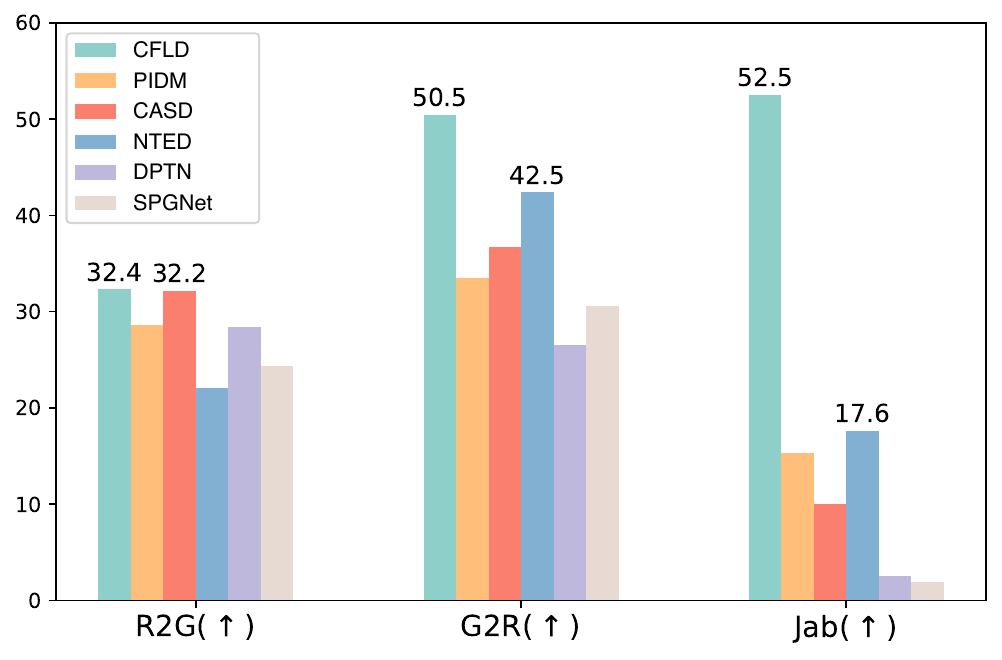}
    \vspace{-0.7cm}
    \caption{
        User study results in terms of R2G, G2R and Jab metrics.
    }
    \vspace{-0.3cm}
    \label{fig:user_study}
\end{figure}

\subsection{User Study}
\label{sec:user_study}

To verify the gap between generated and real images as well as our superiority over the state of the arts, we have recruited over 100 volunteers to perform the following two user studies following PIDM~\cite{pidm}.
\textbf{(1)} For the R2G and G2R metrics, volunteers were asked to discriminate between 30 generated images and 30 real images from the testing set.
Each volunteer could only see the generated images of a specific method, and the pairs of source image and target pose for generation were consistent across methods for a fair comparison.
From the results in \cref{fig:user_study}, chances of a real image being recognized as generated (R2G) are relatively low, and over half of the images we generated are recognized as real (G2R), demonstrating that our method generates more realistic images that are less likely to be judged as fake by humans.
\textbf{(2)} For the Jab metric, each volunteer was asked to choose the best match to the ground truth from the generated images of different methods.
Compared to other methods, our Jab score achieved 52.5 percent, significantly higher (+34.9) than the counterpart in second place, indicating that our method is more preferred and generates better texture details and pose alignment.

\begin{table}[t]
    \centering
    \footnotesize
    \begin{tabular}{*{6}{m{0.9cm}}}
        \toprule
        \makebox[0.9cm][c]{\textbf{Method}} & \makebox[0.9cm][c]{\textbf{Biasing}} & \makebox[0.9cm][c]{\textbf{Trainable}} & \makebox[0.9cm][c]{\textbf{Prompt}} & \makebox[0.9cm][c]{\textbf{LPIPS$\downarrow$}} & \makebox[0.9cm][c]{\textbf{SSIM$\uparrow$}} \\
        \midrule
        \makebox[0.9cm][c]{\ttfamily\detokenize{B1}} & \makebox[0.9cm][c]{} & \makebox[0.9cm][c]{$\bm{K},\bm{V}$} & \makebox[0.9cm][c]{M-S} & \makebox[0.9cm][c]{0.2018} & \makebox[0.9cm][c]{0.6959} \\
        \makebox[0.9cm][c]{\ttfamily\detokenize{B2}} & \makebox[0.9cm][c]{} & \makebox[0.9cm][c]{$\bm{K},\bm{V}$} & \makebox[0.9cm][c]{CLIP} & \makebox[0.9cm][c]{0.2099} & \makebox[0.9cm][c]{0.6944} \\
        \makebox[0.9cm][c]{\ttfamily\detokenize{B3}} & \makebox[0.9cm][c]{} & \makebox[0.9cm][c]{$\bm{K},\bm{V}$} & \makebox[0.9cm][c]{PRD} & \makebox[0.9cm][c]{0.1615} & \makebox[0.9cm][c]{0.7293} \\
        \makebox[0.9cm][c]{\ttfamily\detokenize{B4}} & \makebox[0.9cm][c]{} & \makebox[0.9cm][c]{$\bm{Q},\bm{K},\bm{V}$} & \makebox[0.9cm][c]{PRD} & \makebox[0.9cm][c]{0.1742} & \makebox[0.9cm][c]{0.7198} \\
        \makebox[0.9cm][c]{\ttfamily\detokenize{B5}} & \makebox[0.9cm][c]{$\bm{Q}$} & \makebox[0.9cm][c]{$\bm{K},\bm{V}$} & \makebox[0.9cm][c]{Swin} & \makebox[0.9cm][c]{0.1912} & \makebox[0.9cm][c]{0.7038} \\
        \makebox[0.9cm][c]{Ours} & \makebox[0.9cm][c]{$\bm{Q}$} & \makebox[0.9cm][c]{$\bm{K},\bm{V}$} & \makebox[0.9cm][c]{PRD} & \makebox[0.9cm][c]{0.1519} & \makebox[0.9cm][c]{0.7378} \\
        \bottomrule
    \end{tabular}
    \vspace{-0.1cm}
    \caption{Quantitative results for ablation studies. M-S is short for multi-scale fine-grained appearance features similar to \cite{pidm,pocold}.}
    \label{tab:ablation_study}
\end{table}
\begin{figure}[t]
    \centering
    \includegraphics[width=\linewidth]{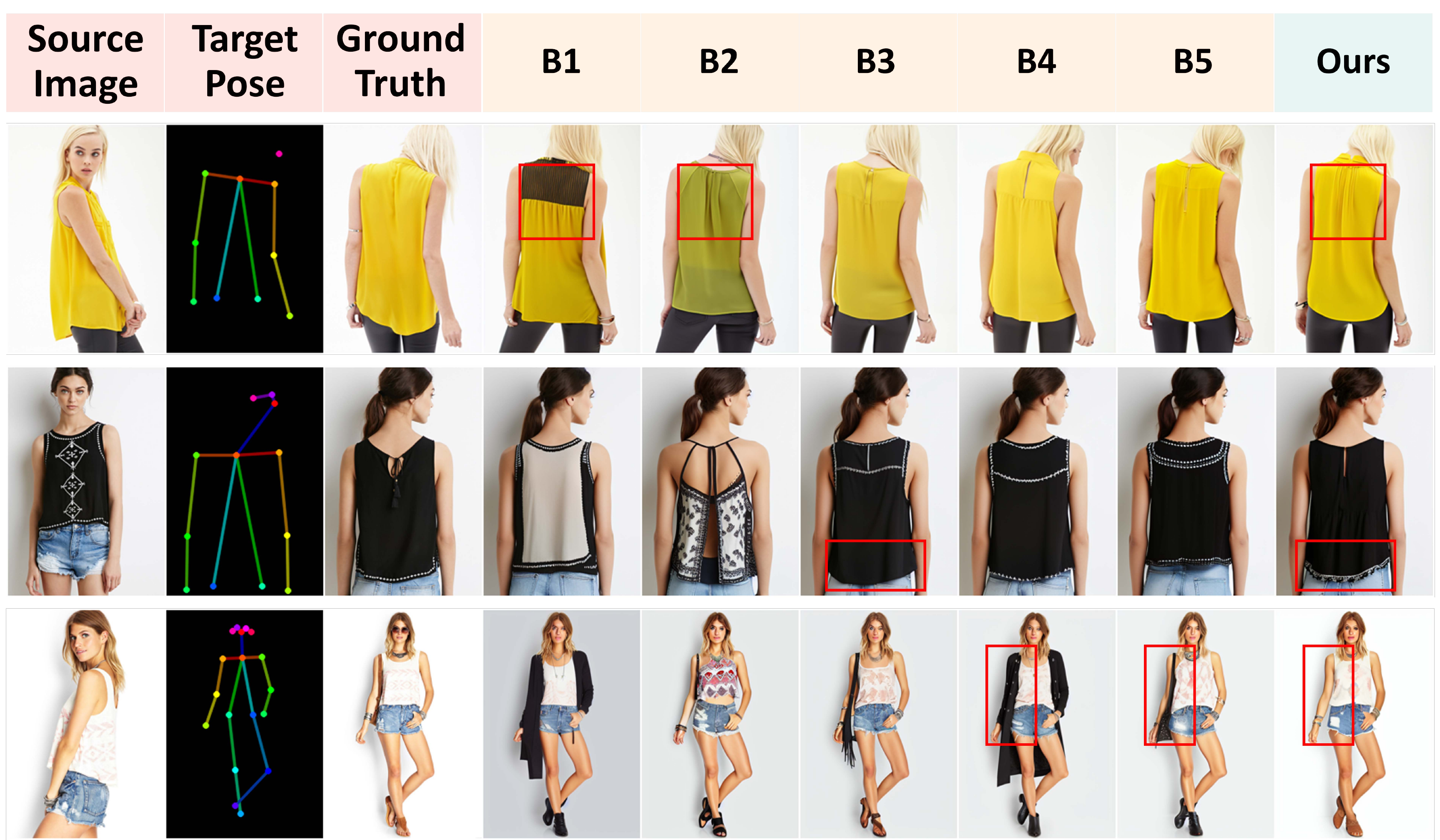}
    \vspace{-0.5cm}
    \caption{
        Qualitative ablation results.
        Our approach has a high-level understanding of the source image rather than forced alignment. 
        It is also less prone to overfitting through the complementary coarse-grained prompts and fine-grained appearance biasing.
    }
    \vspace{-0.3cm}
    \label{fig:ablation}
\end{figure}

\begin{figure*}[t]
    \centering
    \includegraphics[width=\linewidth]{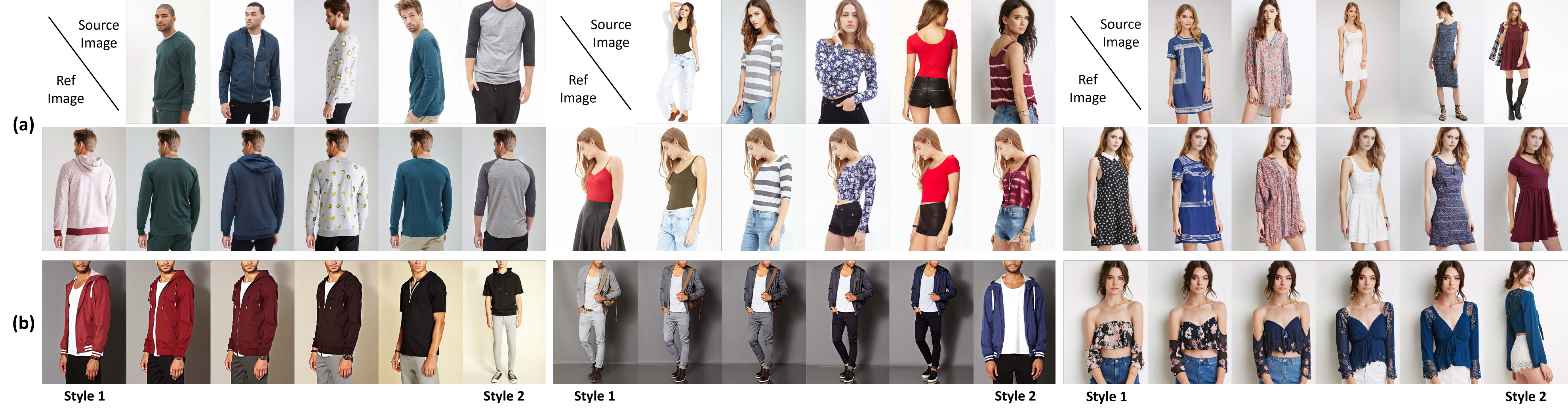}
    \vspace*{-0.7cm}
    \caption{
        (a) Style transfer results of our method.
        The appearance in the reference image can be edited while maintaining the pose and appearance.
        This is achieved by masking out regions of interest in the reference image and requires no additional training.
        (b) The interpolation results show that texture details can be gradually shifted from one style to another in a smooth manner (from Style 1 to 2).
    }
    \vspace*{-0.3cm}
    \label{fig:transfer}
\end{figure*}
\begin{figure}[t]
    \centering
    \includegraphics[width=\linewidth]{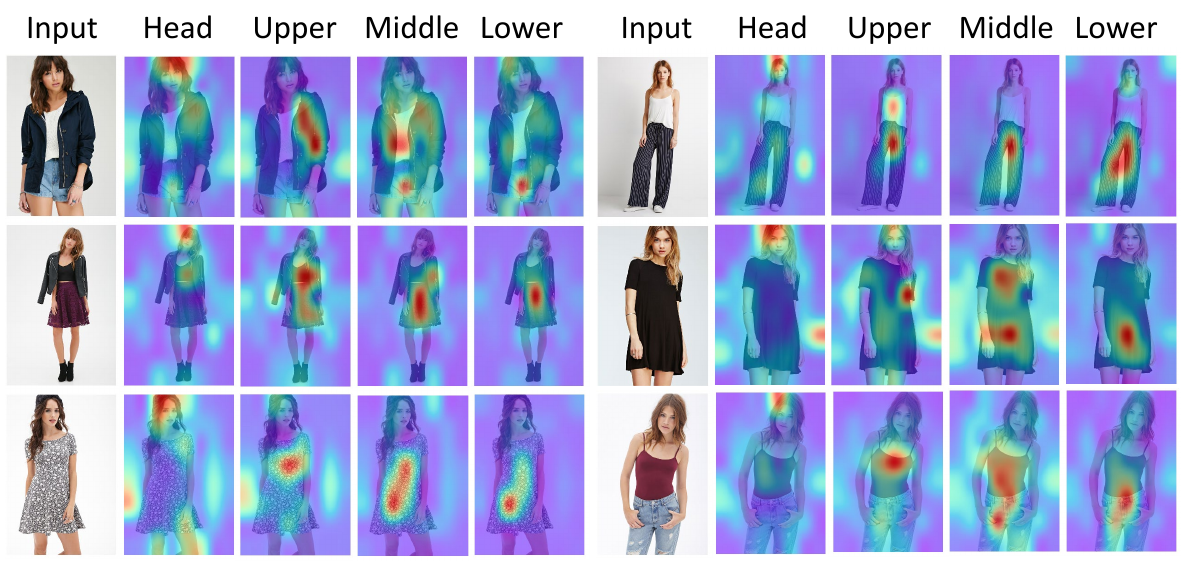}
    \vspace{-0.6cm}
    \caption{
        Visualizing attention maps by different queries of the prompt decoder. 
        The maps are averaged over all attention heads.
    }
    \vspace{-0.3cm}
    \label{fig:visualization}
\end{figure}

\subsection{Ablation Study}
\label{sec:ablation_study}

We perform ablation studies at multiple baselines to compare with our method.
The quantitative results are presented in \cref{tab:ablation_study}.
\verb|B1| is referenced from the other two diffusion-based approaches~\cite{pidm,pocold} that incorporate multi-scale fine-grained appearance features as conditional prompts.
we also experiment with CLIP image encoder~\cite{clip} in \verb|B2| to produce descriptive coarse-grained prompts for source images, which is first explored by an image-editing approach~\cite{paint_by_example} that are also conditioned purely on images.
Together with the qualitative results in \cref{fig:ablation}, we can see that even very simple textures sometimes fail to be preserved, suggesting that these prompts are not compatible with the pre-trained SD model.
To provide coarse-grained features that are more specific to person images, we integrate the Perception-Refined Decoder (PRD) into \verb|B3|.
The reconstruction metrics (i.e., LPIPS and SSIM) in \cref{tab:ablation_study} reveal a significant improvement in the quality of generated images, which validates the effectiveness of our proposed PRD.
While this can be confirmed qualitatively in \cref{fig:ablation}, there is still a lack of textural details as indicated by the red box.
To address this issue, we experiment with training more parameters in the UNet as \verb|B4| and instead observe a decrease in performance.
This implies that the generalization ability of SD model is compromised, which is not our expectation.
Thus we come up with the Hybrid-Granularity Attention (HGA) to bias the queries and achieve state-of-the-art results both quantitatively and qualitatively.
In order to verify whether the source image encoder (i.e., Swin Transformer~\cite{swin_transformer}) is able to learn sufficient information for HGA and give a useful prompt, we abandon the PRD in \verb|B5|.
The qualitative results in \cref{fig:ablation} demonstrate that both \verb|B4| and \verb|B5| are overfitting, only our method circumvents this problem by learning in a coarse to fine-grained manner.

\noindent\textbf{Visualization.}
In \cref{fig:visualization}, we visualize the effectiveness of different queries in $\mathcal{H}_D$.
The attention maps reflect different human body parts of person images captured by learnable queries, which proves that we have a high-level understanding of the source images and thus less prone to overfitting.

\subsection{Appearance Editing}
\noindent\textbf{Style Transfer.}
Our CFLD inherits the strong generation ability of SD model by freezing the vast majority of its parameters.
Thus the style transfer can be achieved simply by masking without additional training.
Specifically, we mark the regions of interest in the reference image $\bm{y}^{ref}$ as a binary mask $\bm{m}$.
During sampling, the noise prediction is decomposed into $\epsilon_t^\prime=\bm{m}\cdot\epsilon_t+(1-\bm{m})\cdot\bm{z}_t^{ref}$, where the $\epsilon_t$ is based on the pose from $\bm{y}^{ref}$ and the appearance from different styles of source images.
Let $\bm{z}_t^{ref}$ be the noisy latent at timestep $t$ mapped from $\bm{z}_0^{ref}=\mathcal{E}(\bm{y}^{ref})$ as in \cref{eq:diffusion}.
From the results in \cref{fig:transfer}(a), our method generates realistic and coherent texture details in the regions of interest.

\noindent\textbf{Style Interpolation.}
Additionally, our CFLD supports arbitrary linear interpolation of both coarse-grained prompts and fine-grained appearance biases.
As shown in \cref{fig:transfer}(b), our generated images are faithfully reproducing different styles with smooth transitions.
\section{Conclusion}
\noindent
This paper presents a novel Coarse-to-Fine Latent Diffusion (CFLD) method for Pose-Guided Person Image Synthesis (PGPIS).
We circumvent the potential overfitting problem by decoupling the fine-grained appearance and pose information controls.
Our proposed Perception-Refined Decoder (PRD) and Hybrid-Granularity Attention module (HGA) enable a high-level semantic understanding of person images, while also preserving texture details through a coarse-to-fine learning curriculum. 
Extensive experiments demonstrate that CFLD outperforms the state of the arts in PGPIS both quantitatively and qualitatively.
Our future work will investigate whether the CFLD can be extended to more downstream tasks that suffer from inferior data like person re-identification~\cite{lu2022improving,yuan2024hap} and domain adaptation~\cite{lu2024mlnet,shen2023collaborative}, since our training paradigm yields both a pre-trained feature network and a powerful generator for augmentation.


\noindent\textbf{Acknowledgments.}
This work was supported in part by the National Natural Science Foundation of China (U22A2095, 62276281).

{
    \small
    \bibliographystyle{ieeenat_fullname}
    \bibliography{main}

\begin{thebibliography}{63}
\providecommand{\natexlab}[1]{#1}
\providecommand{\url}[1]{\texttt{#1}}
\expandafter\ifx\csname urlstyle\endcsname\relax
  \providecommand{\doi}[1]{doi: #1}\else
  \providecommand{\doi}{doi: \begingroup \urlstyle{rm}\Url}\fi

\bibitem[Bhunia et~al.(2023)Bhunia, Khan, Cholakkal, Anwer, Laaksonen, Shah, and Khan]{pidm}
Ankan~Kumar Bhunia, Salman Khan, Hisham Cholakkal, Rao~Muhammad Anwer, Jorma Laaksonen, Mubarak Shah, and Fahad~Shahbaz Khan.
\newblock Person image synthesis via denoising diffusion model.
\newblock In \emph{CVPR}, page 5968–5976, 2023.

\bibitem[Brooks et~al.(2023)Brooks, Holynski, and Efros]{cumulative_cfg}
Tim Brooks, Aleksander Holynski, and Alexei~A Efros.
\newblock Instructpix2pix: Learning to follow image editing instructions.
\newblock In \emph{CVPR}, pages 18392--18402, 2023.

\bibitem[Cao et~al.(2017)Cao, Simon, Wei, and Sheikh]{openpose}
Zhe Cao, Tomas Simon, Shih-En Wei, and Yaser Sheikh.
\newblock Realtime multi-person 2d pose estimation using part affinity fields.
\newblock In \emph{CVPR}, page 7291–7299, 2017.

\bibitem[Deng et~al.(2009)Deng, Dong, Socher, Li, Li, and Fei-Fei]{imagenet}
Jia Deng, Wei Dong, Richard Socher, Li-Jia Li, Kai Li, and Li Fei-Fei.
\newblock Imagenet: A large-scale hierarchical image database.
\newblock In \emph{CVPR}, page 248–255, 2009.

\bibitem[Dhariwal and Nichol(2021)]{dhariwal2021diffusion}
Prafulla Dhariwal and Alexander~Quinn Nichol.
\newblock Diffusion models beat gans on image synthesis.
\newblock In \emph{NeurIPS}, pages 8780--8794, 2021.

\bibitem[Esser and Sutter(2018)]{vunet}
Patrick Esser and Ekaterina Sutter.
\newblock A variational u-net for conditional appearance and shape generation.
\newblock In \emph{CVPR}, page 8857–8866, 2018.

\bibitem[Esser et~al.(2021)Esser, Rombach, and Ommer]{vae}
Patrick Esser, Robin Rombach, and Bjorn Ommer.
\newblock Taming transformers for high-resolution image synthesis.
\newblock In \emph{CVPR}, page 12873–12883, 2021.

\bibitem[G{\"u}ler et~al.(2018)G{\"u}ler, Neverova, and Kokkinos]{densepose}
R{\i}za~Alp G{\"u}ler, Natalia Neverova, and Iasonas Kokkinos.
\newblock Densepose: Dense human pose estimation in the wild.
\newblock In \emph{CVPR}, pages 7297--7306, 2018.

\bibitem[Han et~al.(2023)Han, Zhu, Deng, Song, and Xiang]{pocold}
Xiao Han, Xiatian Zhu, Jiankang Deng, Yi-Zhe Song, and Tao Xiang.
\newblock Controllable person image synthesis with pose-constrained latent diffusion.
\newblock In \emph{ICCV}, page 22768–22777, 2023.

\bibitem[He et~al.(2016)He, Zhang, Ren, and Sun]{resnet}
Kaiming He, Xiangyu Zhang, Shaoqing Ren, and Jian Sun.
\newblock Deep residual learning for image recognition.
\newblock In \emph{CVPR}, page 770–778, 2016.

\bibitem[He et~al.(2022)He, Chen, Xie, Li, Dollár, and Girshick]{mae}
Kaiming He, Xinlei Chen, Saining Xie, Yanghao Li, Piotr Dollár, and Ross Girshick.
\newblock Masked autoencoders are scalable vision learners.
\newblock In \emph{CVPR}, page 16000–16009, 2022.

\bibitem[Hertz et~al.(2022)Hertz, Mokady, Tenenbaum, Aberman, Pritch, and Cohen-Or]{hertz2022prompt}
Amir Hertz, Ron Mokady, Jay Tenenbaum, Kfir Aberman, Yael Pritch, and Daniel Cohen-Or.
\newblock Prompt-to-prompt image editing with cross attention control.
\newblock \emph{arXiv:2208.01626}, 2022.

\bibitem[Heusel et~al.(2017)Heusel, Ramsauer, Unterthiner, Nessler, and Hochreiter]{fid}
Martin Heusel, Hubert Ramsauer, Thomas Unterthiner, Bernhard Nessler, and Sepp Hochreiter.
\newblock Gans trained by a two time-scale update rule converge to a local nash equilibrium.
\newblock In \emph{NeurIPS}, 2017.

\bibitem[Ho and Salimans(2021)]{cfg}
Jonathan Ho and Tim Salimans.
\newblock Classifier-free diffusion guidance.
\newblock In \emph{NeurIPS Workshops}, 2021.

\bibitem[Ho et~al.(2020)Ho, Jain, and Abbeel]{ddpm}
Jonathan Ho, Ajay Jain, and Pieter Abbeel.
\newblock Denoising diffusion probabilistic models.
\newblock In \emph{NeurIPS}, 2020.

\bibitem[Hu et~al.(2023)Hu, Gao, Zhang, Sun, Zhang, and Bo]{animate_anyone}
Liucheng Hu, Xin Gao, Peng Zhang, Ke Sun, Bang Zhang, and Liefeng Bo.
\newblock Animate anyone: Consistent and controllable image-to-video synthesis for character animation.
\newblock \emph{arXiv:2311.17117}, 2023.

\bibitem[Kingma and Ba(2015)]{adam}
Diederik~P. Kingma and Jimmy Ba.
\newblock Adam: {A} method for stochastic optimization.
\newblock In \emph{ICLR}, 2015.

\bibitem[Li et~al.(2019)Li, Huang, and Loy]{diaf}
Yining Li, Chen Huang, and Chen~Change Loy.
\newblock Dense intrinsic appearance flow for human pose transfer.
\newblock In \emph{CVPR}, page 3693–3702, 2019.

\bibitem[Liu and Chilton(2022)]{prompt1}
Vivian Liu and Lydia~B Chilton.
\newblock Design guidelines for prompt engineering text-to-image generative models.
\newblock In \emph{CHI}, pages 1--23, 2022.

\bibitem[Liu et~al.(2019)Liu, Piao, Min, Luo, Ma, and Gao]{liquid}
Wen Liu, Zhixin Piao, Jie Min, Wenhan Luo, Lin Ma, and Shenghua Gao.
\newblock Liquid warping gan: A unified framework for human motion imitation, appearance transfer and novel view synthesis.
\newblock In \emph{ICCV}, pages 5904--5913, 2019.

\bibitem[Liu et~al.(2016)Liu, Luo, Qiu, Wang, and Tang]{deepfashion}
Ziwei Liu, Ping Luo, Shi Qiu, Xiaogang Wang, and Xiaoou Tang.
\newblock Deepfashion: Powering robust clothes recognition and retrieval with rich annotations.
\newblock In \emph{CVPR}, page 1096–1104, 2016.

\bibitem[Liu et~al.(2021)Liu, Lin, Cao, Hu, Wei, Zhang, Lin, and Guo]{swin_transformer}
Ze Liu, Yutong Lin, Yue Cao, Han Hu, Yixuan Wei, Zheng Zhang, Stephen Lin, and Baining Guo.
\newblock Swin transformer: Hierarchical vision transformer using shifted windows.
\newblock In \emph{ICCV}, page 10012–10022, 2021.

\bibitem[Lu et~al.(2022)Lu, Zhang, Lin, Ma, Xie, and Lai]{lu2022improving}
Yanzuo Lu, Manlin Zhang, Yiqi Lin, Andy~J Ma, Xiaohua Xie, and Jianhuang Lai.
\newblock Improving pre-trained masked autoencoder via locality enhancement for person re-identification.
\newblock In \emph{Chinese Conference on Pattern Recognition and Computer Vision (PRCV)}, pages 509--521. Springer, 2022.

\bibitem[Lu et~al.(2024)Lu, Shen, Ma, Xie, and Lai]{lu2024mlnet}
Yanzuo Lu, Meng Shen, Andy~J Ma, Xiaohua Xie, and Jian-Huang Lai.
\newblock Mlnet: Mutual learning network with neighborhood invariance for universal domain adaptation.
\newblock In \emph{AAAI}, 2024.

\bibitem[Lv et~al.(2021)Lv, Li, Li, Li, Lin, He, and Zuo]{spgnet}
Zhengyao Lv, Xiaoming Li, Xin Li, Fu Li, Tianwei Lin, Dongliang He, and Wangmeng Zuo.
\newblock Learning semantic person image generation by region-adaptive normalization.
\newblock In \emph{CVPR}, page 10806–10815, 2021.

\bibitem[Ma et~al.(2017)Ma, Jia, Sun, Schiele, Tuytelaars, and Gool]{pg2}
Liqian Ma, Xu Jia, Qianru Sun, B. Schiele, T. Tuytelaars, and L. Gool.
\newblock Pose guided person image generation.
\newblock In \emph{NeurIPS}, 2017.

\bibitem[Ma et~al.(2018)Ma, Sun, Georgoulis, Van~Gool, Schiele, and Fritz]{disentangled}
Liqian Ma, Qianru Sun, Stamatios Georgoulis, Luc Van~Gool, Bernt Schiele, and Mario Fritz.
\newblock Disentangled person image generation.
\newblock In \emph{CVPR}, pages 99--108, 2018.

\bibitem[Men et~al.(2020)Men, Mao, Jiang, Ma, and Lian]{adgan}
Yifang Men, Yiming Mao, Yuning Jiang, Wei-Ying Ma, and Zhouhui Lian.
\newblock Controllable person image synthesis with attribute-decomposed gan.
\newblock In \emph{CVPR}, page 5084–5093, 2020.

\bibitem[Mou et~al.(2023)Mou, Wang, Xie, Zhang, Qi, Shan, and Qie]{t2i_adapter}
Chong Mou, Xintao Wang, Liangbin Xie, Jing Zhang, Zhongang Qi, Ying Shan, and Xiaohu Qie.
\newblock T2i-adapter: Learning adapters to dig out more controllable ability for text-to-image diffusion models.
\newblock \emph{arXiv:2302.08453}, 2023.

\bibitem[Paszke et~al.(2019)Paszke, Gross, Massa, Lerer, Bradbury, Chanan, Killeen, Lin, Gimelshein, Antiga, et~al.]{pytorch}
Adam Paszke, Sam Gross, Francisco Massa, Adam Lerer, James Bradbury, Gregory Chanan, Trevor Killeen, Zeming Lin, Natalia Gimelshein, Luca Antiga, et~al.
\newblock Pytorch: An imperative style, high-performance deep learning library.
\newblock \emph{NeurIPS}, 32, 2019.

\bibitem[Pavlichenko and Ustalov(2023)]{prompt2}
Nikita Pavlichenko and Dmitry Ustalov.
\newblock Best prompts for text-to-image models and how to find them.
\newblock In \emph{SIGIR}, pages 2067--2071, 2023.

\bibitem[Radford et~al.(2021)Radford, Kim, Hallacy, Ramesh, Goh, Agarwal, Sastry, Askell, Mishkin, Clark, Krueger, and Sutskever]{clip}
Alec Radford, Jong~Wook Kim, Chris Hallacy, Aditya Ramesh, Gabriel Goh, Sandhini Agarwal, Girish Sastry, Amanda Askell, Pamela Mishkin, Jack Clark, Gretchen Krueger, and Ilya Sutskever.
\newblock Learning transferable visual models from natural language supervision.
\newblock In \emph{ICML}, page 8748–8763, 2021.

\bibitem[Ren et~al.(2020)Ren, Yu, Chen, Li, and Li]{gfla}
Yurui Ren, Xiaoming Yu, Junming Chen, Thomas~H. Li, and Ge Li.
\newblock Deep image spatial transformation for person image generation.
\newblock In \emph{CVPR}, page 7690–7699, 2020.

\bibitem[Ren et~al.(2022)Ren, Fan, Li, Liu, and Li]{nted}
Yurui Ren, Xiaoqing Fan, Ge Li, Shan Liu, and Thomas~H. Li.
\newblock Neural texture extraction and distribution for controllable person image synthesis.
\newblock In \emph{CVPR}, page 13535–13544, 2022.

\bibitem[Rombach et~al.(2022)Rombach, Blattmann, Lorenz, Esser, and Ommer]{stable_diffusion}
Robin Rombach, Andreas Blattmann, Dominik Lorenz, Patrick Esser, and Bjorn Ommer.
\newblock High-resolution image synthesis with latent diffusion models.
\newblock In \emph{CVPR}, page 10684–10695, 2022.

\bibitem[Ronneberger et~al.(2015)Ronneberger, Fischer, and Brox]{unet}
Olaf Ronneberger, Philipp Fischer, and Thomas Brox.
\newblock U-net: Convolutional networks for biomedical image segmentation.
\newblock In \emph{MICCAI}, page 234–241, 2015.

\bibitem[Salimans et~al.(2016)Salimans, Goodfellow, Zaremba, Cheung, Radford, and Chen]{inception}
Tim Salimans, Ian Goodfellow, Wojciech Zaremba, Vicki Cheung, Alec Radford, and Xi Chen.
\newblock Improved techniques for training gans.
\newblock \emph{NeurIPS}, 29, 2016.

\bibitem[Sarkar et~al.(2021)Sarkar, Golyanik, Liu, and Theobalt]{uvmap}
Kripasindhu Sarkar, Vladislav Golyanik, Lingjie Liu, and Christian Theobalt.
\newblock Style and pose control for image synthesis of humans from a single monocular view.
\newblock \emph{arXiv:2102.11263}, 2021.

\bibitem[Shen et~al.(2024)Shen, Ye, Zhang, Wang, Han, and Yang]{pcdms}
Fei Shen, Hu Ye, Jun Zhang, Cong Wang, Xiao Han, and Wei Yang.
\newblock Advancing pose-guided image synthesis with progressive conditional diffusion models.
\newblock In \emph{ICLR}, 2024.

\bibitem[Shen et~al.(2023)Shen, Lu, Hu, and Ma]{shen2023collaborative}
Meng Shen, Yanzuo Lu, Yanxu Hu, and Andy~J Ma.
\newblock Collaborative learning of diverse experts for source-free universal domain adaptation.
\newblock In \emph{ACM MM}, pages 2054--2065, 2023.

\bibitem[Siarohin et~al.(2018)Siarohin, Sangineto, Lathuiliere, and Sebe]{pose_gan}
Aliaksandr Siarohin, Enver Sangineto, Stephane Lathuiliere, and Nicu Sebe.
\newblock Deformable gans for pose-based human image generation.
\newblock In \emph{CVPR}, page 3408–3416, 2018.

\bibitem[Song et~al.(2021{\natexlab{a}})Song, Meng, and Ermon]{ddim}
Jiaming Song, Chenlin Meng, and Stefano Ermon.
\newblock Denoising diffusion implicit models.
\newblock In \emph{ICLR}, 2021{\natexlab{a}}.

\bibitem[Song et~al.(2021{\natexlab{b}})Song, Sohl{-}Dickstein, Kingma, Kumar, Ermon, and Poole]{sde}
Yang Song, Jascha Sohl{-}Dickstein, Diederik~P. Kingma, Abhishek Kumar, Stefano Ermon, and Ben Poole.
\newblock Score-based generative modeling through stochastic differential equations.
\newblock In \emph{ICLR}, 2021{\natexlab{b}}.

\bibitem[Tang et~al.(2020)Tang, Bai, Zhang, Torr, and Sebe]{xinggan}
Hao Tang, Song Bai, Li Zhang, Philip H.~S. Torr, and Nicu Sebe.
\newblock Xinggan for person image generation.
\newblock In \emph{ECCV}, page 717–734, 2020.

\bibitem[Tumanyan et~al.(2023)Tumanyan, Geyer, Bagon, and Dekel]{tumanyan2023plug}
Narek Tumanyan, Michal Geyer, Shai Bagon, and Tali Dekel.
\newblock Plug-and-play diffusion features for text-driven image-to-image translation.
\newblock In \emph{CVPR}, pages 1921--1930, 2023.

\bibitem[Vaserstein(1969)]{wasserstein}
Leonid~Nisonovich Vaserstein.
\newblock Markov processes over denumerable products of spaces, describing large systems of automata.
\newblock \emph{Problemy Peredachi Informatsii}, 5\penalty0 (3):\penalty0 64--72, 1969.

\bibitem[Vaswani et~al.(2017)Vaswani, Shazeer, Parmar, Uszkoreit, Jones, Gomez, Kaiser, and Polosukhin]{transformer}
Ashish Vaswani, Noam Shazeer, Niki Parmar, Jakob Uszkoreit, Llion Jones, Aidan~N. Gomez, Lukasz Kaiser, and Illia Polosukhin.
\newblock Attention is all you need.
\newblock In \emph{NeurIPS}, 2017.

\bibitem[von Platen et~al.(2022)von Platen, Patil, Lozhkov, Cuenca, Lambert, Rasul, Davaadorj, and Wolf]{diffusers}
Patrick von Platen, Suraj Patil, Anton Lozhkov, Pedro Cuenca, Nathan Lambert, Kashif Rasul, Mishig Davaadorj, and Thomas Wolf.
\newblock Diffusers: State-of-the-art diffusion models.
\newblock \url{https://github.com/huggingface/diffusers}, 2022.

\bibitem[Wang et~al.(2004)Wang, Bovik, Sheikh, and Simoncelli]{ssim}
Zhou Wang, Alan~C. Bovik, Hamid~R. Sheikh, and Eero~P. Simoncelli.
\newblock Image quality assessment: From error visibility to structural similarity.
\newblock \emph{TIP}, 13\penalty0 (4):\penalty0 600–612, 2004.

\bibitem[Xu et~al.(2023)Xu, Zhang, Liew, Yan, Liu, Zhang, Feng, and Shou]{magicanimate}
Zhongcong Xu, Jianfeng Zhang, Jianfeng Liew, Hanshu Yan, Jia-Wei Liu, Chenxu Zhang, Jiashi Feng, and Mike~Zheng Shou.
\newblock Magicanimate: Temporally consistent human image animation using diffusion model.
\newblock \emph{arXiv:2311.16498}, 2023.

\bibitem[Yang et~al.(2023)Yang, Gu, Zhang, Zhang, Chen, Sun, Chen, and Wen]{paint_by_example}
Binxin Yang, Shuyang Gu, Bo Zhang, Ting Zhang, Xuejin Chen, Xiaoyan Sun, Dong Chen, and Fang Wen.
\newblock Paint by example: Exemplar-based image editing with diffusion models.
\newblock In \emph{CVPR}, page 18381–18391, 2023.

\bibitem[Ye et~al.(2023)Ye, Zhang, Liu, Han, and Yang]{ip_adapter}
Hu Ye, Jun Zhang, Siyi Liu, Xiao Han, and Wei Yang.
\newblock Ip-adapter: Text compatible image prompt adapter for text-to-image diffusion models.
\newblock \emph{arXiv:2308.06721}, 2023.

\bibitem[Yuan et~al.(2024)Yuan, Zhang, Zhou, Wang, Qiu, Shao, Zhang, Long, Kuang, Yao, et~al.]{yuan2024hap}
Junkun Yuan, Xinyu Zhang, Hao Zhou, Jian Wang, Zhongwei Qiu, Zhiyin Shao, Shaofeng Zhang, Sifan Long, Kun Kuang, Kun Yao, et~al.
\newblock Hap: Structure-aware masked image modeling for human-centric perception.
\newblock \emph{NeurIPS}, 36, 2024.

\bibitem[Zhang et~al.(2021)Zhang, Li, Lai, and Yang]{pise}
Jinsong Zhang, Kun Li, Yu-Kun Lai, and Jingyu Yang.
\newblock Pise: Person image synthesis and editing with decoupled gan.
\newblock In \emph{CVPR}, page 7982–7990, 2021.

\bibitem[Zhang et~al.(2023)Zhang, Rao, and Agrawala]{controlnet}
Lvmin Zhang, Anyi Rao, and Maneesh Agrawala.
\newblock Adding conditional control to text-to-image diffusion models.
\newblock In \emph{ICCV}, page 3836–3847, 2023.

\bibitem[Zhang et~al.(2022)Zhang, Yang, Lai, and Xie]{dptn}
Pengze Zhang, Lingxiao Yang, Jianhuang Lai, and Xiaohua Xie.
\newblock Exploring dual-task correlation for pose guided person image generation.
\newblock In \emph{CVPR}, page 7713–7722, 2022.

\bibitem[Zhang et~al.(2018)Zhang, Isola, Efros, Shechtman, and Wang]{lpips}
Richard Zhang, Phillip Isola, Alexei~A. Efros, Eli Shechtman, and Oliver Wang.
\newblock The unreasonable effectiveness of deep features as a perceptual metric.
\newblock In \emph{CVPR}, page 586–595, 2018.

\bibitem[Zheng et~al.(2015)Zheng, Shen, Tian, Wang, Wang, and Tian]{market_1501}
Liang Zheng, Liyue Shen, Lu Tian, Shengjin Wang, Jingdong Wang, and Qi Tian.
\newblock Scalable person re-identification: A benchmark.
\newblock In \emph{ICCV}, page 1116–1124, 2015.

\bibitem[Zhou et~al.(2022{\natexlab{a}})Zhou, Yang, Loy, and Liu]{cocoop}
Kaiyang Zhou, Jingkang Yang, Chen~Change Loy, and Ziwei Liu.
\newblock Conditional prompt learning for vision-language models.
\newblock In \emph{CVPR}, page 16816–16825, 2022{\natexlab{a}}.

\bibitem[Zhou et~al.(2022{\natexlab{b}})Zhou, Yang, Loy, and Liu]{coop}
Kaiyang Zhou, Jingkang Yang, Chen~Change Loy, and Ziwei Liu.
\newblock Learning to prompt for vision-language models.
\newblock \emph{IJCV}, 130\penalty0 (9):\penalty0 2337–2348, 2022{\natexlab{b}}.

\bibitem[Zhou et~al.(2021)Zhou, Zhang, Zhang, Zhang, Bao, Chen, Zhang, and Wen]{cocosnetv2}
Xingran Zhou, Bo Zhang, Ting Zhang, Pan Zhang, Jianmin Bao, Dong Chen, Zhongfei Zhang, and Fang Wen.
\newblock Cocosnet v2: Full-resolution correspondence learning for image translation.
\newblock In \emph{CVPR}, page 11465–11475, 2021.

\bibitem[Zhou et~al.(2022{\natexlab{c}})Zhou, Yin, Chen, Sun, Gao, and Li]{casd}
Xinyue Zhou, M. Yin, Xinyuan Chen, Li Sun, Changxin Gao, and Qingli Li.
\newblock Cross attention based style distribution for controllable person image synthesis.
\newblock In \emph{ECCV}, page 161–178, 2022{\natexlab{c}}.

\bibitem[Zhu et~al.(2019)Zhu, Huang, Shi, Yu, Wang, and Bai]{patn}
Zhen Zhu, Tengteng Huang, Baoguang Shi, Miao Yu, Bofei Wang, and Xiang Bai.
\newblock Progressive pose attention transfer for person image generation.
\newblock In \emph{CVPR}, page 2347–2356, 2019.

\end{thebibliography}
}


\clearpage
\noindent\textbf{Evaluation on Market-1501.}
Since none of the diffusion-based methods including PIDM~\cite{pidm}, PoCoLD~\cite{pocold} and concurrent PCDMs~\cite{pcdms} have released generated images or checkpoints on Market-1501~\cite{market_1501}, we make fair comparisons with available GAN-based methods in \cref{tab:market}.
From these results, CFLD still outperforms across different metrics faithfully, which validates our robustness.

\noindent\textbf{Ablation on classifier-free strategy.}
In the \cref{tab:cfg} we vary the choices of Eq.(7) on DeepFashion~\cite{deepfashion}.
The results show that appropriate reinforcement of both appearance and pose information (i.e., increase guidance weights) can effectively improve the quality of generated images.

\begin{table}[t]
    \centering
    \footnotesize
    \begin{tabular}{m{1.8cm}m{1cm}*{4}{m{0.7cm}}}
        \toprule
        \makebox[1.8cm][c]{\textbf{Method}} & \makebox[1cm][l]{\textbf{Venue}} &\makebox[0.7cm][c]{\textbf{FID$\downarrow$}} & \makebox[0.7cm][c]{\textbf{LPIPS$\downarrow$}} & \makebox[0.7cm][c]{\textbf{SSIM$\uparrow$}} & \makebox[0.7cm][c]{\textbf{PSNR$\uparrow$}}\\
        \midrule

        \rowcolor[gray]{0.9} \multicolumn{6}{l}{\textit{GAN-based Methods}} \\


        \makebox[1.8cm][l]{GFLA} & \makebox[1cm][r]{CVPR 20'} & \makebox[0.7cm][r]{19.740} & \makebox[0.7cm][c]{0.2815} & \makebox[0.7cm][c]{0.2808} & \makebox[0.7cm][c]{14.337} \\

        \makebox[1.8cm][l]{XingGAN} & \makebox[1cm][r]{ECCV 20'} & \makebox[0.7cm][r]{22.520} & \makebox[0.7cm][c]{0.3058} & \makebox[0.7cm][c]{0.3044} & \makebox[0.7cm][c]{14.446} \\
        
        \makebox[1.8cm][l]{SPGNet} & \makebox[1cm][r]{CVPR 21'} & \makebox[0.7cm][r]{23.057} & \makebox[0.7cm][c]{0.2777} & \makebox[0.7cm][c]{\underline{0.3139}} & \makebox[0.7cm][c]{14.489} \\
        
        \makebox[1.8cm][l]{DPTN} & \makebox[1cm][r]{CVPR 22'} & \makebox[0.7cm][r]{\underline{18.995}} & \makebox[0.7cm][c]{\underline{0.2711}} & \makebox[0.7cm][c]{0.2854} & \makebox[0.7cm][c]{\underline{14.521}} \\
        
        \midrule

        \rowcolor[gray]{0.9} \multicolumn{6}{l}{\textit{Diffusion-based Methods}} \\

        \multicolumn{2}{l}{\textbf{CFLD (Ours)}} & \makebox[0.7cm][r]{\textbf{11.972}} & \makebox[0.7cm][c]{\textbf{0.2636}} & \makebox[0.7cm][c]{\textbf{0.3173}} & \makebox[0.7cm][c]{\textbf{14.861}} \\
        
        \multicolumn{2}{l}{\textcolor{gray}{VAE Reconstructed}} & \makebox[0.7cm][r]{\textcolor{gray}{6.028}} & \makebox[0.7cm][c]{\textcolor{gray}{0.0164}} & \makebox[0.7cm][c]{\textcolor{gray}{0.9883}} & \makebox[0.7cm][c]{\textcolor{gray}{36.625}} \\
        
        \multicolumn{2}{l}{\textcolor{gray}{Ground Truth}} & \makebox[0.7cm][r]{\textcolor{gray}{4.845}} & \makebox[0.7cm][c]{\textcolor{gray}{0.0000}} & \makebox[0.7cm][c]{\textcolor{gray}{1.0000}} & \makebox[0.7cm][c]{\textcolor{gray}{$+\infty$}} \\
        
        \bottomrule
    \end{tabular}
    \caption{Quantitative comparisons with the state of the arts on Market-1501~\cite{market_1501} dataset.}
    \label{tab:market}
\end{table}

\begin{figure*}[t]
    \centering
    \includegraphics[width=\linewidth]{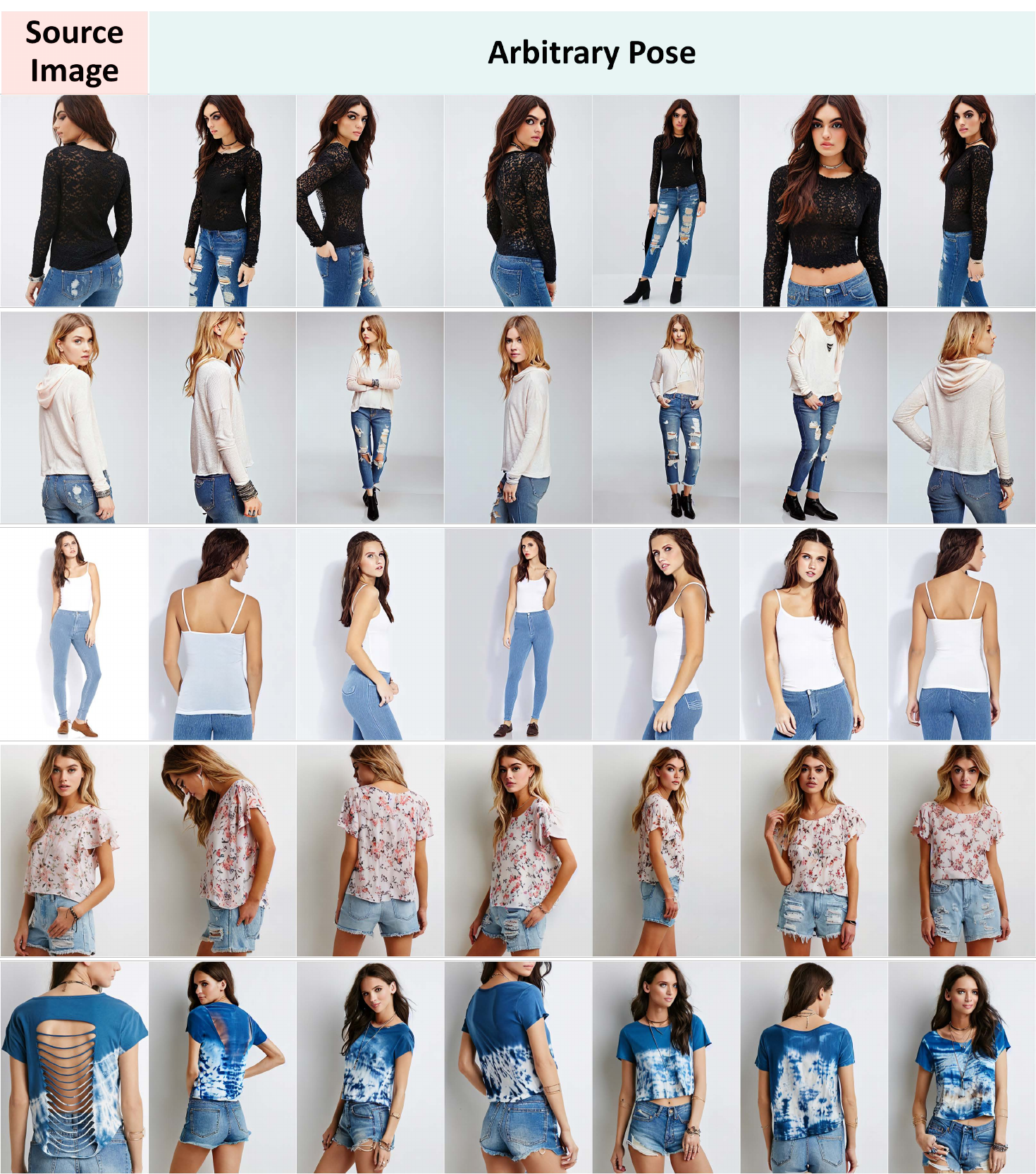}
    \caption{
        Additional results on arbitrary poses from the test set.
    }
    \label{fig:arbitrary1}
\end{figure*}

\begin{figure*}[t]
    \centering
    \includegraphics[width=\linewidth]{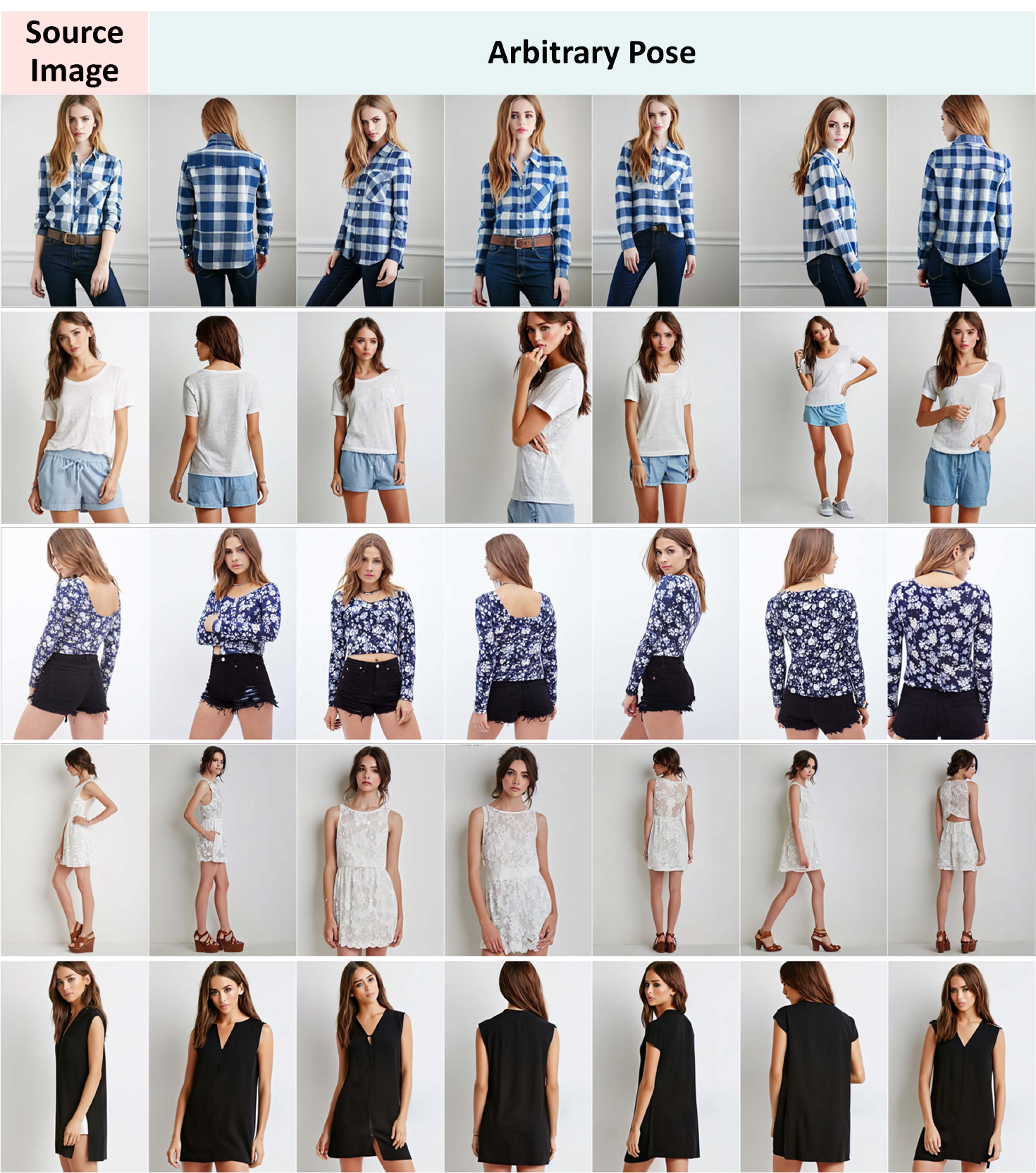}
    \caption{
        Additional results on arbitrary poses from the test set.
    }
    \label{fig:arbitrary2}
\end{figure*}

\begin{figure*}[t]
    \centering
    \includegraphics[width=\linewidth]{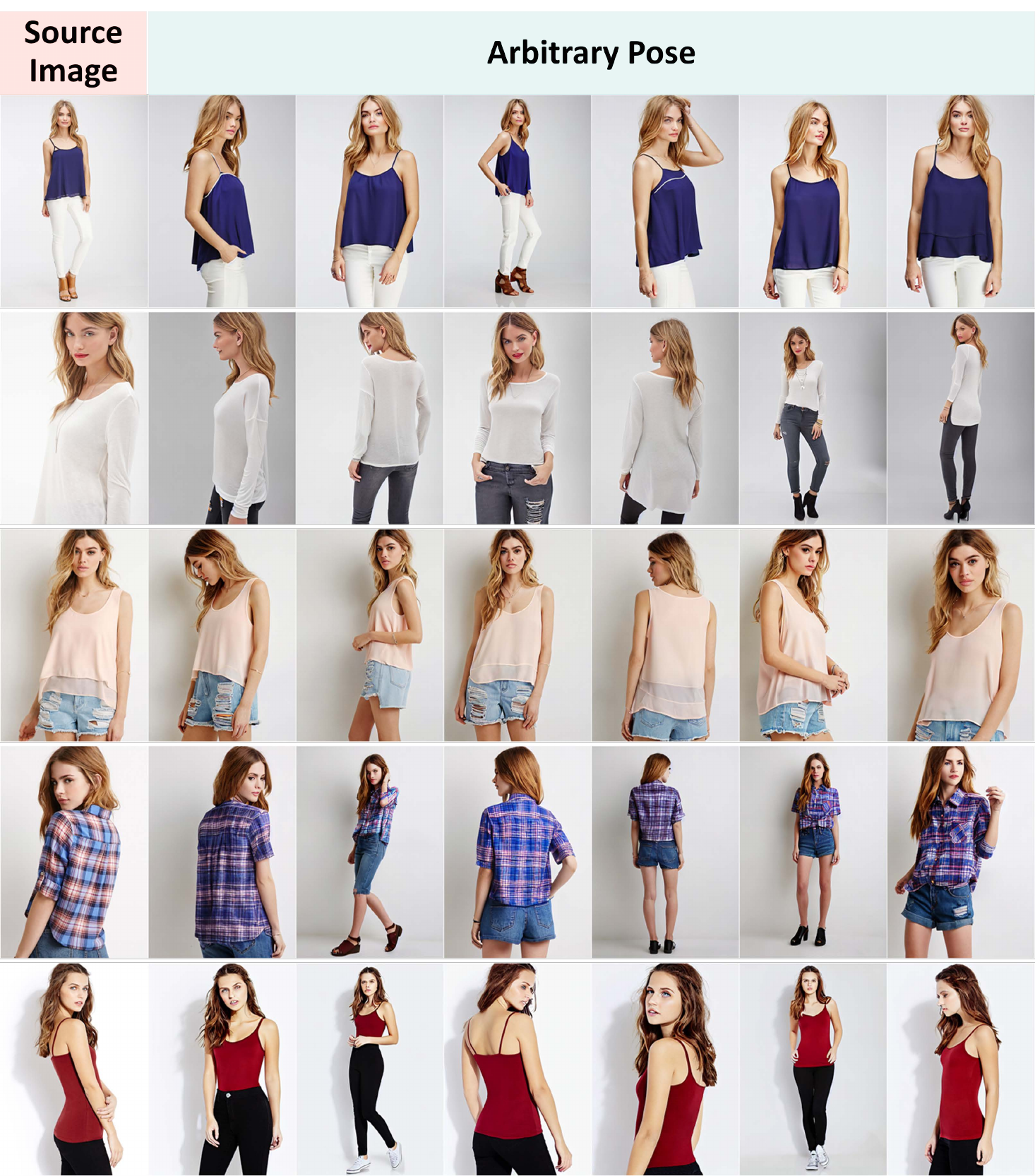}
    \caption{
        Additional results on arbitrary poses from the test set.
    }
    \label{fig:arbitrary3}
\end{figure*}

\noindent \textbf{{Additional qualitative results.}}
To further evaluate the generalization ability of our method, we generate person images at arbitrary poses randomly selected from the test set following in \cref{fig:arbitrary1,fig:arbitrary2,fig:arbitrary3}.
The results show that our method consistently generate high-quality person images while preserving the appearance in the source image.
Even if the target pose differs significantly from the source image, or if invisible areas of the source image are required, the generated images are still free of distortion.
With the guidance of coarse-grained prompts, our method has a high-level understanding and does not suffer from overfitting such as forcing the texture details of the source image to be aligned.
On this basis, our embedded hybrid-granularity attention only supplements the necessary fine-grained appearance features, thus enabling more realistic and natural textures.

\noindent\textbf{More discussion of over-fitting and biasing.}
Our observation is that previous diffusion-based methods would fit the spatially convolutional features of source image into noisy sample directly.
But this doesn't make sense in practice, because the texture details of source image probably shouldn't be present in the same position of target sample, especially in the exaggerated pose transition case.
Since the model is actually performing copy-and-paste, the generations are distorted and unnatural, which we call this phenomenon \textbf{overfitting} and lack of generalization ability.

To circumvent it, we made three efforts: 
1) We introduce pre-trained text-to-image diffusion as foundation model to improve generalization ability since it has been exposed to billions of image-text pairs.
This empowers the model to speculate on some regions of the target pose that are not visible in the source image.
2) Note that textual description for PGPIS task is not available.
To promote efficient fine-tuning without loss of generalization, we freeze most parameters in diffusion model (98.8\%) and thus forcing the proposed PRD to learn coarse-grained semantics just as what the CLIP text encoder provide.
3) To decouple the fine-grained appearance and pose information as opposed to previous approaches, we endeavour to encode the multi-scale convolutional features as bias terms into cross-attention.
The multi-scale \textbf{biasing} would be necessary since the coarse-grained prompts learned solely by the PRD may lack the preservation of texture details, given that the conditional prompt is the same for each scale in U-Net blocks.
We leave the biased queries ($\bm{Q}$ in Eq.(4)) untrained and adopt zero convolution designs both in order to reduce the learning velocity of the HGA module thereby promoting a coarse-to-fine appearance control as stated in manuscript.

\begin{table}[t]
    \centering
    \footnotesize
    \begin{tabular}{m{1.5cm}m{0.5cm}m{0.5cm}*{4}{m{0.7cm}}}
        \toprule
        \makebox[1.5cm][c]{\textbf{Strategy}} & 
        \makebox[0.5cm][c]{\textbf{$\bm{w}_{\text{pose}}$}} & 
        \makebox[0.5cm][c]{\textbf{$\bm{w}_{\text{app}}$}} & 
        \makebox[0.7cm][c]{\textbf{FID$\downarrow$}} & \makebox[0.7cm][c]{\textbf{LPIPS$\downarrow$}} & \makebox[0.7cm][c]{\textbf{SSIM$\uparrow$}} & \makebox[0.7cm][c]{\textbf{PSNR$\uparrow$}}\\
        \midrule

        \makebox[1.5cm][c]{disabled} & \makebox[0.5cm][c]{1.0} & \makebox[0.5cm][c]{1.0} & \makebox[0.7cm][r]{8.143} & \makebox[0.7cm][c]{0.2000} & \makebox[0.7cm][c]{0.7055} & \makebox[0.7cm][c]{15.753} \\
        \makebox[1.5cm][c]{appearance only} & \makebox[0.5cm][c]{1.0} & \makebox[0.5cm][c]{2.0} & \makebox[0.7cm][r]{8.334} & \makebox[0.7cm][c]{0.1921} & \makebox[0.7cm][c]{0.7131} & \makebox[0.7cm][c]{16.429} \\
        \makebox[1.5cm][c]{pose only} & \makebox[0.5cm][c]{2.0} & \makebox[0.5cm][c]{1.0} & \makebox[0.7cm][r]{7.580} & \makebox[0.7cm][c]{\underline{0.1770}} & \makebox[0.7cm][c]{\underline{0.7256}} & \makebox[0.7cm][c]{17.611} \\
        \makebox[1.5cm][c]{\textbf{both}} & \makebox[0.5cm][c]{\textbf{2.0}} & \makebox[0.5cm][c]{\textbf{2.0}} & \makebox[0.7cm][r]{\textbf{6.804}} & \makebox[0.7cm][c]{\textbf{0.1519}} & \makebox[0.7cm][c]{\textbf{0.7378}} & \makebox[0.7cm][c]{\textbf{18.235}} \\
        \makebox[1.5cm][c]{both} & \makebox[0.5cm][c]{3.0} & \makebox[0.5cm][c]{3.0} & \makebox[0.7cm][r]{\underline{7.423}} & \makebox[0.7cm][c]{0.1746} & \makebox[0.7cm][c]{0.7250} & \makebox[0.7cm][c]{\underline{17.706}} \\
        
        \bottomrule
    \end{tabular}
    \caption{Ablation on classifier-free strategy.}
    \label{tab:cfg}
\end{table}

\end{document}